%% file: main.tex
\documentclass[10pt,twocolumn,letterpaper]{article}

\usepackage[pagenumbers]{cvpr} %

\usepackage{graphicx}
\usepackage{amsmath}
\usepackage{amssymb}
\usepackage{booktabs}

\usepackage{tabularx}       %
\usepackage{multirow}  
\usepackage{graphicx}
\usepackage{xspace}
\usepackage{capt-of,etoolbox}

\usepackage{pgfplots}
\pgfplotsset{compat=1.5, every axis/.append style={font=\small, /pgf/number format/1000 sep={}}}
\usepackage{tikz}
\usetikzlibrary{quotes, angles, calc, 3d, shapes, intersections, plotmarks}
\usepgfplotslibrary{statistics, polar, groupplots}
\pgfplotsset{every mark/.append style={solid},} %
\usepackage{tikzscale}

\input{preamble}
\definecolor{cvprblue}{rgb}{0.21,0.49,0.74}
\usepackage[pagebackref,breaklinks,colorlinks,allcolors=cvprblue]{hyperref}

\usepackage[capitalize]{cleveref}
\crefname{section}{Sec.}{Secs.}
\Crefname{section}{Section}{Sections}
\Crefname{table}{Table}{Tables}
\crefname{table}{Tab.}{Tabs.}

\input{macros}

\newcolumntype{C}{>{\centering\arraybackslash}X}

\def\paperID{13} %
\def\confName{CVPR}
\def\confYear{2025}

\graphicspath{{./figures/}}

\newcommand\splashfigure{%
    \centering
    \includegraphics[height=5cm]{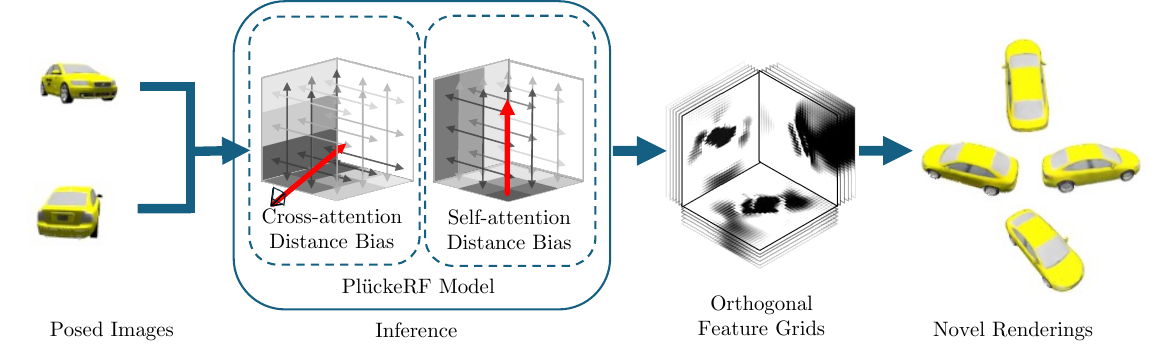}
    \captionof{figure}{\method: A feed-forward method that directly predicts a 3D representation from posed input images. The model uses a line--line distance biased attention mechanism
    that helps the model interpret which parts of the images are important for constructing specific areas of the 3D output. The resulting 3D representation is structured as three orthogonal feature grids, enabling efficient novel view synthesis.}
    \label{fig:splash-image} %
}

\makeatletter
\apptocmd\@maketitle{{\splashfigure{}\par\vspace{12pt}}}{}{}
\makeatother

\title{\method: A Line-based 3D Representation for Few-view Reconstruction}

\author{Sam Bahrami\\
The Australian National University\\
Canberra, Australia\\
{\tt\small sam.bahrami@anu.edu.au}
\and
Dylan Campbell\\
The Australian National University\\
Canberra, Australia\\
{\tt\small dylan.campbell@anu.edu.au}
}

\begin{document}
\maketitle

\begin{abstract}

Feed-forward 3D reconstruction methods aim to predict the 3D structure of a scene directly from input images, providing a faster alternative to per-scene optimization approaches. 
Significant progress has been made in single-view and few-view reconstruction using learned priors that infer object shape and appearance, even for unobserved regions. 
However, there is substantial potential to enhance these methods by better leveraging information from multiple views when available.
To address this, we propose a few-view reconstruction model that more effectively harnesses multi-view information. 
Our approach introduces a simple mechanism that connects the 3D representation with pixel rays from the input views, allowing for preferential sharing of information between nearby 3D locations and between 3D locations and nearby pixel rays. 
We achieve this by defining the 3D representation as a set of structured, feature-augmented lines---the \method representation. 
Using this representation, we demonstrate improvements in reconstruction quality over the equivalent triplane representation and state-of-the-art feedforward reconstruction methods.

\end{abstract}

\section{Introduction}
\label{sec:intro}
3D reconstruction is the task of estimating the shape and appearance of objects or scenes from data. 
This typically involves creating a 3D model from a set of 2D images, which may include additional information such as camera pose or depth maps.
With enough images, methods like Neural Radiance Fields (NeRF) \cite{mildenhall_nerf_2020} and more recently Gaussian Splatting \cite{kerbl_3d_2023} allows one to reconstruct a 3D scene through novel view synthesis (NVS).
The challenge comes when there is limited data, which is the case for few-view and single-view 3D reconstruction tasks. 

Initial neural representations for NVS were based on fully-connected multi-layer perceptron (MLP) models \cite{mildenhall_nerf_2020}.
Since then interpretable neural representations have been developed, such as voxel feature grids \cite{liu_neural_2020}, hashing \cite{muller_instant_2022}, hybrid implicit--explicit triplane representations \cite{chan_eg3d_2021, chen_tensorf_2022}, and Gaussian mixtures \cite{kerbl_3d_2023}. 
Each of these representations presents its own trade-offs in terms of computational efficiency, memory usage, and rendering quality.

Voxel grids and triplane representations are learnable 3D structures that encode information in the form of feature grids and orthogonal feature planes, respectively. 
To infer a novel view, a point in 3D space is queried through this representation to obtain its corresponding feature encoding, which is processed by a small neural network to determine the pixel’s color and density. 
Unlike MLP representations, voxel grids and triplanes offer an inherently interpretable structure, enabling the application of useful inductive biases, such as convolutions that share information between neighboring points. 
This structured nature allows voxel grids and triplanes to be inferred through image-to-image methods, such as encoder--decoder style feed-forward neural networks \cite{hong_lrm_2023, szymanowicz_viewset_2023}. 

Several single-view 3D reconstruction methods have been adapted to work with additional, but still few, views \cite{szymanowicz_splatter_2024, yu_pixelnerf_2021}. 
This is often referred to as few-view reconstruction.
Often this adaptation is done \textit{without} providing additional context about the relationships between the views.
For example, by inferring an explicit representation from each individual view and merging the individual outputs \cite{zhang2024gslrmlargereconstructionmodel}, simply concatenating additional image and pose encodings to a models input \cite{li2023instant3dfasttextto3dsparseview, jin_lvsm_2024}, or averaging feature representations across individual image and pose encodings \cite{yu_pixelnerf_2021}.

Previous work has shown that incorporating an image pixel ray-to-ray distance bias can improve the quality of few-view 3D reconstructions \cite{venkat_geometry-biased_2023}. 
We define a distance measure that quantifies the proximity between a camera ray for an image pixel, and a ray orthogonal to a pixel of our orthogonal feature grid representation---the \method representation. 
Explicitly we define this as a line-to-line distance measure, by converting these rays to origin invariant line representations.
We use these distances to bias the attention mechanism in a feed-forward transformer model, encouraging information sharing between lines that are close or intersecting while penalizing sharing between distant lines. 
The intuition for this is that image regions corresponding to specific areas of the 3D representation should exert a stronger influence on those areas within the model.
A high level summary of our approach can be seen in \cref{fig:splash-image}.
By incorporating additional contextual information from a few supplementary views (two views in our experiments), our approach enhances the accuracy and realism of the novel views inferred from our 3D neural field.
Our contributions are
\begin{enumerate}
    \item the \method representation that facilitates geometrically-meaningful sharing of information within the 3D representation and between it and the available input images; and
    \item a simple few-view reconstruction model that uses this internal representation to predict a 3D neural field in a single forward pass.
\end{enumerate}

\section{Related Work}

Commonly, 3D reconstructions are represented using point clouds \cite{wu_pq-net_2020}, meshes \cite{wang_pixel2mesh_2018}, neural representations \cite{mildenhall_nerf_2020, chen_tensorf_2022} or Gaussian splats \cite{kerbl_3d_2023}. 
Models designed for single or few-view reconstruction typically learn a prior from a dataset, and with sufficiently large datasets, this prior can enable robust generalization. 

\paragraph{Neural 3D Reconstruction and Representations.}
A neural radiance field (NeRF) \cite{mildenhall_nerf_2020} is a machine learning model that learns the mapping between a 3D point and view direction with a corresponding color and density. 
The original NeRF approach employs a multi-layer perceptron (MLP) neural network to define an implicit function which takes a point and a ray direction, and returns a color and density.
To render a posed image, a ray is cast through each pixel. 
Multiple positions along each ray are then queried using the model, informed by the camera's pose, and volumetric rendering is applied to compute the final pixel color.
Hashing methods have addressed some of the drawbacks of the original NeRF, by using a multi-resolution hash table encoding that efficiently represents spatial data and leveraging hierarchical representations to quickly capture both global and fine-grained details in the scene \cite{muller_instant_2022}.

Subsequent works implemented alternative structured representations such as pixel-aligned Gaussian mixtures \cite{szymanowicz_splatter_2024, charatan_pixelsplat_2023, smart2024splatt3r} based on Gaussian splats \cite{kerbl_3d_2023}, voxel grids \cite{liu_neural_2020, szymanowicz_viewset_2023}, and triplane feature grids \cite{chan_eg3d_2021, chen_tensorf_2022, gu_nerfdiff_2023}. 
These structured representations offer memory-efficient 3D representations of the scene, at a trade-off with the representation resolution. 
We use a triplane-like representation, as it provides greater resolution with a smaller memory footprint compared to a voxel representation, and we can make use of the structured representation to guide our \method model. 

\paragraph{Single-View 3D Object Reconstruction. }
3D representations such as point clouds \cite{wu_pq-net_2020, wiles2020synsin}, meshes \cite{xu2024instantmesh}, signed distance fields \cite{park_deepsdf_2019}, NeRFs \cite{muller_autorf_2022, jang_codenerf_2021, yu_pixelnerf_2021, hong_lrm_2023}, and Gaussian Splats \cite{szymanowicz_splatter_2024} have dominated learning-based single-view reconstruction approaches.
These methods learn a prior representation across a dataset, which generalizes to new objects or scenes \cite{muller_autorf_2022, szymanowicz_splatter_2024, chan_eg3d_2021, venkat_geometry-biased_2023, szymanowicz_flash3d_2024}.
Initial single-view NeRF methods parameterize the MLP through latent representations from the source image \cite{yu_pixelnerf_2021, jang_codenerf_2021}.
Later works used structured NeRF representations such as voxel feature grids \cite{szymanowicz_viewset_2023}, and hybrid implicit--explicit triplane representations \cite{chan_eg3d_2021, hong_lrm_2023}. 
With triplane representations, large feed-forward reconstruction models have become a popular method for single and few-view 3D reconstruction \cite{hong_lrm_2023}. 
These models use large scale datasets (\eg, 800k+ objects \cite{deitke_objaverse_2022}) which helps generalize across categories.

Another approach has been to use generative models to convert the single--view reconstruction to a multi-view reconstruction problem. 
These approaches may fine-tune their 3D representation using generated multi--view instances \cite{liu2023zero} or use estimated images from other views for guidance in image--to--3D reconstruction \cite{melas-kyriazi_realfusion_2023, gu_nerfdiff_2023}.

Most recently pixel aligned Gaussian mixture representations have also shown strong performance in single--view reconstruction, taking advantage of the very fast rendering to speed up training, requiring less compute resources to get results comparable to those using the triplane representations \cite{szymanowicz_splatter_2024, szymanowicz_flash3d_2024}.

Our \method approach is based on a feed-forward reconstruction model \cite{hong_lrm_2023, szymanowicz_splatter_2024}. 
These models are very simple, often comprised of purely transformer layers, and by using an orthogonal feature grid representation, like a triplane, our model can infer the complete feature representation in a forward pass.

\paragraph{Few-View 3D Object Reconstruction. }
Few-view often means a number of input views between two \cite{yu_pixelnerf_2021, szymanowicz_splatter_2024} and six \cite{venkat_geometry-biased_2023}. 
Utilizing multiple views provides more context to the model in two ways, first by reducing the amount of unobserved object surface area, and second by making it possible for a model to triangulate 3D surface geometry.
Many of the previous single view reconstruction tasks have been adapted to use multiple views.
Some of these methods simply merge representations from each individual view \cite{szymanowicz_splatter_2024, zhang2024gslrmlargereconstructionmodel, yu_pixelnerf_2021}, or provide the additional image and camera pose to their model as concatenated image encoding tokens \cite{li2023instant3dfasttextto3dsparseview}.

Other approaches leverage geometric relationships between images to enhance 3D reconstruction. 
Even without known camera poses, it is possible to predict the relative 3D poses of unposed 2D images and estimate the scaled depth of associated points in each image \cite{wang_dust3r_2024, wang_vggsfm_2023, leroy_mast3r_2024, wang_cut3r_2025}. 
This serves as a valuable starting point for reconstructing scenes from image pairs \cite{smart2024splatt3r}.
Another approach jointly optimizes object shape and camera poses from few-view inputs, reducing reliance on precise pose initialization and yielding consistent 3D reconstructions \cite{yang_fvor_2022, jiang_forge_2024}.

When camera poses are known, this information can be utilized to determine which parts of the reconstruction each individual image should focus on \cite{venkat_geometry-biased_2023}.
3D representation-free novel view synthesis models have emerged that directly map sparse views to high-quality novel images without explicit 3D representations \cite{jin_lvsm_2024}.
Many of these multi-view reconstruction methods use \plucker coordinates as a positional encoding of their inputs \cite{sitzmann_lfn_2021, chen_ray-conditioning_2023, jin_lvsm_2024}.

In contrast, our work utilizes the pose information from the few views, and geometric information from our structured feature representation, using a \plucker coordinate representation for both.
Pixels in our input images can be attended to based on their relationship with the underlying 3D representation---our \method representation.

\section{\method Few-view Reconstruction}
\label{sec:method}

The goal of this work is to render novel viewpoints of previously unseen objects from a few posed images. 
To do so, we train a transformer-based neural network to infer three orthogonal feature grids from the provided input views.
We begin by providing background information on \plucker coordinates and the feature grid representation. 
Next, we provide a high level overview of our model, followed by details on our 2D image representation and our geometric feature grid representation, explaining how they interact within our transformer attention mechanism. 
We conclude with detail on our pretrained image encoder and training information. 

\begin{figure*}[!t]
  \centering
  \includegraphics[trim=0 0 0 0, height=6.2cm]{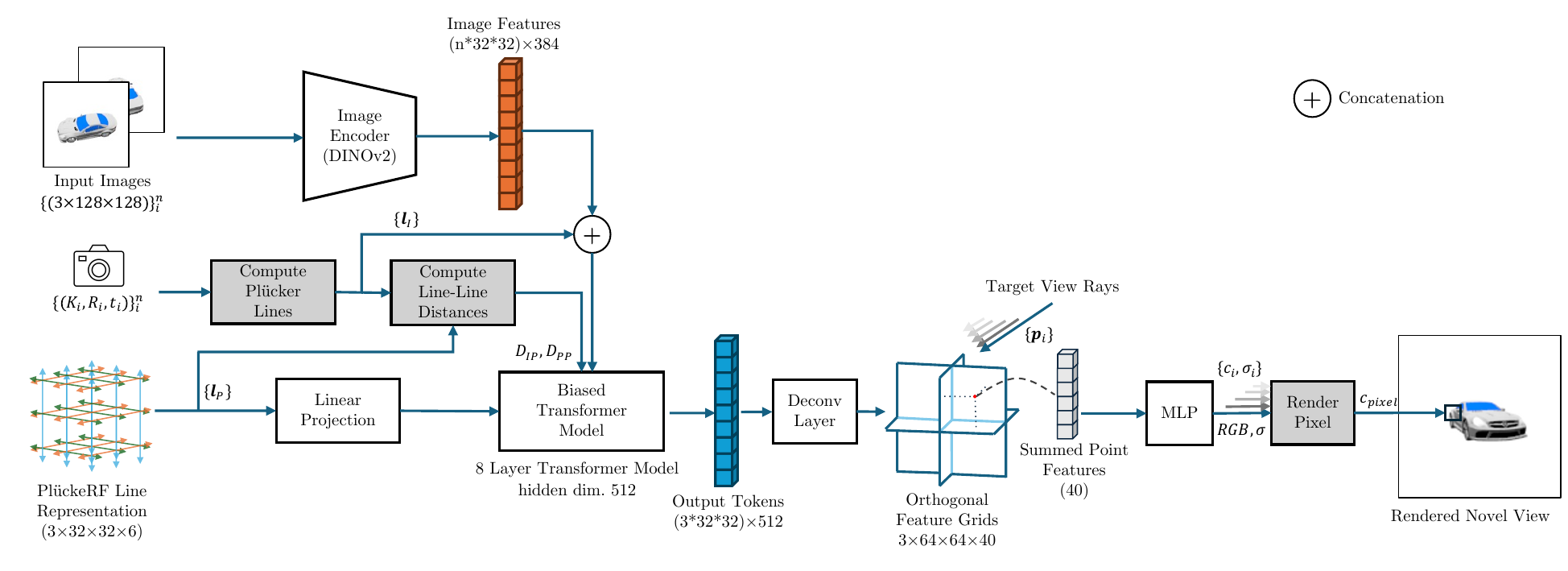}
  \caption{Flowchart for the \method model.
  We use a pre-trained vision transformer (DINOv2 \cite{oquab_dinov2_2023}) to encode the input images (\cref{sec:image-encoder}).
  We take the camera information from the images and create rays corresponding with the image patches from our image encoder, which we convert to \plucker coordinates (\cref{sec:image-plucker-coordinates}).
  Our final input is the set of lines corresponding to the pixels in the orthogonal feature grid representation, which we call the \method line representation (\cref{sec:plucker-representation}).
  We calculate the distance between these \plucker coordinates and those from our input images, and apply these as an attention bias in our transformer attention blocks (\cref{sec:3dattn}).
  The output of the transformer is reshaped to three orthogonal feature planes, which we use for novel view synthesis (\cref{sec:triplane-nerf}).
  Our network is trained end-to-end with image reconstruction losses (\cref{sec:training-objective}).}
  \label{fig:architecture-diagram}
\end{figure*}

\subsection{Preliminaries}
\label{sec:preliminaries}
\paragraph{Orthogonal Feature Grid Representation.}
We use three orthogonal, axis-aligned feature grids that are aligned with the world coordinate system to model the opacity and radiance features at every 3D point. 
This is also known as a triplane representation \cite{chan_eg3d_2021}.
Concretely, this representation is defined as a set of three orthogonal axis-aligned feature grids $\mathcal{T} = \{\bT_{xy}, \bT_{yz}, \bT_{zx}\}$ where $\bT_{xy}, \bT_{yz}, \bT_{zx} \in \mathbb{R}^{M \times M \times d_T}$ with $M$ being the square grid width and height respectively, and $d_T$ is the feature dimension.
The grid covers the dilated object bounding box $[-1, 1]^3$, which defines the 3D space the object exists in. 
For details on how this representation is used to render novel views, see \cref{sec:triplane-nerf}.

\paragraph{\plucker Coordinates.}
\plucker coordinates are a line representation $\bl = (\bd, \bem) \in \mathbb{P}^5$ (projective 5-space) that may be computed from a ray with origin $\bo \in \reals^3$ and direction $\bd \in \mathbb{S}^2$ by taking the cross product \cite{plucker_1865}:
\begin{equation}
\bl = (\bd, \bem) = (\bd, \bo \times \bd)
\label{eq:plucker-coordinates}
\end{equation}
They are invariant to the choice of origin along the ray and are homogeneous coordinates for the line, that is, $(\lambda\bd, \lambda\bem)$ for $\lambda \neq 0$ represents the same line.
We fix $\lambda=1$.
Given two lines represented by \plucker coordinates $\bl_1 = (\bd_1, \bem_1)$ and $\bl_2 = (\bd_2, \bem_2)$,
the closest distance between the lines is given by
\begin{equation}
d(\bl_1, \bl_2) \!=\!\!
\begin{cases} 
| \bd_1\transpose \bem_2 \!+\! \bd_2\transpose \bem_1 | / \|\bd_1 \!\times\! \bd_2\|_2
& \text{if } \bd_1 \!\times\! \bd_2 \neq 0 \\
\|\bd_1 \!\times\! (\bem_1 \!-\! (\bd_1\transpose \bd_2)\bem_2)\|_2 & \text{otherwise.}
\end{cases}
\label{eq:distance}
\end{equation}
When $\bd_1 \times \bd_2 \neq 0$, the lines are not parallel, allowing us to compute the shortest distance between skew lines. 
In the other case, the lines are parallel and the distance is calculated by projecting the moment vectors onto a plane perpendicular to their common direction.

\subsection{Images-to-Feature-Grid Transformer Decoder}
\label{sec:architecture}

A visual summary of our method can be seen in \cref{fig:architecture-diagram}.
Our method infers three orthogonal feature grids from the provided input views.
We use a transformer decoder architecture with cross-attention and self-attention layers within each transformer block.
The initial input to our model is a linear projection of the \method line parameters (\cref{sec:plucker-representation}) to the transformer hidden dimension $d_D$. 
This corresponds with the shape of the three feature grid representation, which are internally represented as flattened feature grids of $3N^2 \times d_D$ dimension.

Our few-view images are applied to the transformer model through the cross-attention layers, where each image is pre-processed into patch-wise image features from a pretrained image encoder, detailed in \cref{sec:image-encoder}.
\plucker coordinates for each patch are computed from the camera information and concatenated to each pixel patch in its feature dimension, as done in other works as a positional encoding \cite{jin_lvsm_2024, sitzmann_lfn_2021, chen_ray-conditioning_2023}.
In addition to patch-wise image features, the DINOv2 encoder provides an additional CLS token.
Since the CLS token does not correspond with any particular part of the image, we set the distance between it and all of the \method lines to 0---the CLS token should incur no distance-based attention penalty.
The set of $n$ input images are then concatenated together.

The resulting output of our transformer model is reshaped to our three orthogonal feature grids. 
We process this through a deconvolution layer to upsample the representation, \ie, from ${(3 \times N \times N \times d_D)}$ to ${(3 \times M \times M \times d_T)}$, where $d_T$ is our orthogonal grids feature dimension, and $M = 2N$.
These feature grids are then used to render the input views and unseen views of our object, using standard NeRF volumetric rendering. 
These views are compared with ground-truth images to calculate a training loss and optimize the model architecture. 
The model hence learns a prior for the training dataset.

\subsection{Image \plucker Coordinates}
\label{sec:image-plucker-coordinates}

Our model processes a set of $n$ input images $\{I_i\}_{i=1}^{n}$.
These images are pre-processed into feature vectors of pixel patches using a pretrained DINOv2 \cite{oquab_dinov2_2023} image encoder (see \cref{sec:image-encoder}).
We utilize \plucker coordinates to represent lines from the camera center passing through the middle of the pixel patches in the encoded image.
These rays are generated using the camera intrinsics $\bK \in \mathbb{R}^{3 \times 3}$ and extrinsics $\bR \in \mathbb{R}^{3 \times 3}$, $\bt \in \mathbb{R}^3$ relative to the first input camera for each image in our set.

Let $\mathbf{u} = (u, v, 1)\transpose$ be the homogeneous coordinates of a pixel center in the image.
The corresponding ray direction $\mathbf{d}$ in the world coordinate system is given by $\mathbf{d} = (\bR \bK^{-1} \mathbf{u}) / \|\bK^{-1} \mathbf{u}\|$.
The origin of the ray $\mathbf{o}$ is the camera center, given by the translation vector $\bo = \bt$.
We convert these rays $\mathbf{r} = (\mathbf{o}, \mathbf{d})$ to \plucker coordinates $\bl$ following \cref{eq:plucker-coordinates}.
This results in a set of \plucker coordinates corresponding to each pixel patch of each input image, denoted as $\{\bl_{I_i}\}_{i=1}^{n}$ for the lines from each image $i$ in our image set.

The conversion to \plucker coordinates abstracts away the specific location of the camera center, focusing instead on the direction of the rays.
\plucker coordinates represent lines based solely on their direction and moment, independent of any specific point along the line, thereby ensuring that translations along the ray do not change the encoding.
This invariance to the camera's position along the line is advantageous in our model, as moving the camera forward or backward along a pixel ray should not significantly affect the color or geometry associated with that particular pixel.
As a result, the representation is robust to variations in the camera's position along the view direction.
The invariance also simplifies the model's learning process by eliminating the need to identify different inputs (camera positions) as being the same (defining the same ray).

\subsection{The \method Representation}
\label{sec:plucker-representation}

Our model infers three orthogonal feature grids which are used for neural rendering.
We define rays orthogonal to each grid plane through each pixel center of our orthogonal feature grids, at the resolution of our model's internal representation, \ie, $3 \times N \times N$ rays.
We convert these rays to lines following \cref{eq:plucker-coordinates}, and denote this set of lines as $\{\bl_{P_i}\}_{i=1}^{3N^2}$---our \method line representation.
We use these to establish a direct geometric relationship between the feature grid features and the input image features through a line--line distance-biased attention.
The input tokens to our biased transformer are instantiated as a linear projection of $\{\bl_{P_i}\}_{i=1}^{3N^2}$ to the model hidden dimension $d_D$.

\subsection{Line--line Distance-biased Attention}
\label{sec:3dattn}

\begin{figure}[!t]\centering
    \begin{subfigure}[]{0.45\linewidth}\centering
        \includegraphics[width=\linewidth]{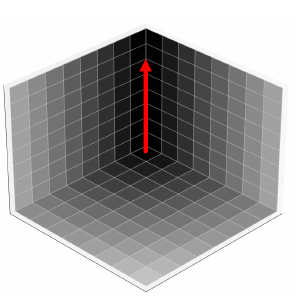}
        \caption{Line--line distances for self-attention.}
        \label{fig:self_attention_lines}
    \end{subfigure}\hfill 
    \begin{subfigure}[]{0.45\linewidth}\centering
        \includegraphics[width=\linewidth]{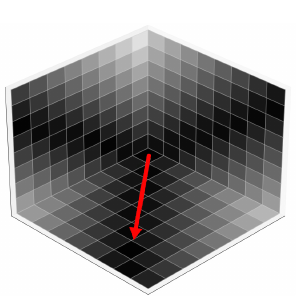}
        \caption{Line--line distances for cross-attention.}
        \label{fig:cross-attention-lines}
    \end{subfigure}
    \caption{A visual representation of the \plucker coordinate distance between a line from our transformer keys (represented by the red arrow), and the lines corresponding to the queries---the \method line representation. The shading on the squares on each axis-aligned plane represents the distance between $\{\bl_{P_i}\}_{i=1}^{3N^2}$ and the red arrow. In practice, those with a greater distance (lighter squares) will be penalized in attending to this feature.
    \subref{fig:self_attention_lines}~The self-attention case, where the red arrow is one of the lines $\{\bl_{P_i}\}_{i=1}^{3N^2}$ from the 3D \method representation.
    \subref{fig:cross-attention-lines}~The cross-attention case, where the red arrow corresponds to a ray from an input camera passing through an image pixel patch.}
    \label{fig:plucker-distance}
\end{figure}

We add a distance bias to the attention mechanism, based on the line-to-line distance between each corresponding image patch $\{\bl_{I_i}\}_{i=1}^{n}$ for our $n$ images, and the lines of our \method representation $\{\bl_{P_i}\}_{i=1}^{3N^2}$.
The intuition is that (a) distant lines should have low attention weights, and hence have a high distance penalty, (b) nearby lines should have a smaller penalty, and hence higher attention weights, and (c) intersecting lines should not be penalized at all (see \cref{fig:plucker-distance}). 
This encourages information to be shared between neighboring and intersecting \method pixels and reduces sharing between unrelated parts of the representation.
We observe that the latter might be sub-optimal in the presence of symmetries and so provide a mechanism to attenuate this bias if it is unhelpful (see $\gamma$, below).

The standard scaled dot product attention mechanism in a transformer cross-attention layer is calculated as follows, using the query $\bQ$, key $\bK$, and value $\bV$ matrices,
\begin{equation}
    \softmax \left({\frac{\bQ\bK\transpose}{\sqrt {d_I}}}\right)  \bV, 
\end{equation}
where the keys and values $\bK, \bV \in \mathbb{R}^{nE^2 \times d_I}$ come from our encoded image features, for $n$ views, image encodings of resolution $E$, and feature dimension $d_I$.
The query matrix $\bQ \in \mathbb{R}^{3N^2 \times d_I}$ originates from a linear projection of the \method line representation—the 3D representation derived from our orthogonal feature grids.

In our model, we bias the attention matrix with the distance between our lines representing the keys $\mathbf{l}_k \in \{\bl_{I_i}\}_{i=1}^{n}$ and our line-based 3D representation representing the queries $\mathbf{l}_q \in \{\bl_{P_i}\}_{i=1}^{3N^2}$,
\begin{equation}
\label{eq:distance-weighting}
    \softmax \left({\frac{\bQ\bK\transpose}{\sqrt {d_I}}} - \gamma d(\mathbf{l}_q, \mathbf{l}_k)\right)  \bV, %
\end{equation}
where the distance is calculated using \Cref{eq:distance}, and $\gamma \in \reals_+$ is a learnable parameter controlling the relative importance of the distance bias.
This provides the model an inductive bias to attend to image features which intersect with areas of the feature grid representation, see \cref{fig:cross-attention-lines}.

In the self-attention layers, we compare each line in $\{\bl_{P_i}\}_{i=1}^{3N^2}$ with all other lines in the set for its distance calculation, for every pixel in our feature grids. 
This provides an inductive bias similar to that of a convolutional layer in a neural network, where neighboring parts of the image share information, but extends this to a 3D locality bias \textit{while only using 2D data structures} (see \cref{fig:self_attention_lines}).
This contrasts with 3D convolutional layers, which have a locality bias, but require processing a volumetric data structure, which is less memory and compute efficient.
Unlike convolutional layers, our approach retains the flexibility of transformers, which can attend across the entire feature map to learn global and local relationships.
Additionally, the learnable parameter $\gamma$ allows the bias to be ignored if it is not useful.

We denote these pairwise distance matrices as $\bD_{IP}$ for the cross-attention and $\bD_{PP}$ for the self-attention cases.

\subsection{Image Encoder}
\label{sec:image-encoder}

Given RGB images as an input, our method first applies a pre-trained vision transformer (ViT) \cite{dosovitskiy2020image} to encode the image to patch-wise feature tokens.
We initialize this as a pretrained DINOv2 \cite{oquab_dinov2_2023} model.
The model uses $14 \times 14$ pixel image patches. %

\subsection{Neural Rendering}
\label{sec:triplane-nerf}

We follow the triplane NeRF formulation which was introduced by Chan \etal \cite{chan_eg3d_2021}.
We use an MLP to predict a color and density from the point features queried from the orthogonal feature grid representation $\mathcal{T} = \{\bT_{xy}, \bT_{yz}, \bT_{zx}\}$.

We compute rays corresponding to each target pixel for our target view given its camera information.
We sample $r$ points $\{\mathbf{p}_i\}_{i=1}^{r}$ along the ray to determine the feature representation, and repeat this ray marching process for every pixel in our rendered view.

For neural rendering, each point $\mathbf{p} \in \mathbb{R}^3$ is projected onto each plane to obtain three feature vectors, $\bbf_{xy}, \bbf_{yz}, \bbf_{zx}$. 
For example, feature vector $\bbf_{xy} \in \reals^{d_T}$ is obtained by projecting $\mathbf{p}$ onto the plane and bilinearly interpolating the grid feature of $\bT_{xy}$ at that location. 
That is $\bbf_{xy} = \text{interp}(\text{proj}_{xy}(\mathbf{p}), \bT_{xy})$ for $\text{proj}_{xy}(\mathbf{p}) = [p_x, p_y]$ and $\text{interp}$ being the bilinear interpolation function.
This is similarly done to obtain $\bbf_{yz}, \bbf_{zx}$.

These features are summed together and processed by a small decoder MLP network to interpret the features as color (RGB) and density $\sigma$.
Our features $\mathbf{f_p}$ at point $\mathbf{p} = (p_x, p_y, p_z)$ are defined by $\mathbf{f_p} = \bbf_{xy} + \bbf_{yz} + \bbf_{zx},$ where $\mathbf{f_p} \in \reals^{d_T}$. 
These are processed by an MLP to predict the color and density at that point,
\begin{align}
(c_i, \sigma_i) &= f(\bp_i) = \text{MLP}(\mathbf{f_p}).
\end{align}
These quantities are rendered into individual pixels and complete RGB images at a given resolution using (neural) volume rendering. 

\subsection{Training Objective}
\label{sec:training-objective}

We render complete views using few ($n$) posed input images and guide the process with $k$ additional ground-truth views around the object. 
During training, we render all $n + k$ views and compare these with their ground-truth views.
To evaluate the quality of the reconstruction, we apply image reconstruction losses. Concretely our reconstruction loss is:
\begin{equation}
    \mathcal{L}(\mathbf{x}) = \frac{1}{n + k} \sum_{v=1}^{n + k} \left( \mathcal{L}_{\text{MSE}}(\hat{x}_v, x_v) + \alpha \mathcal{L}_{\text{LPIPS}}(\hat{x}_v, x_v) \right),
\end{equation}
where $\hat{x}_v$ is our rendered image, $x_v$ is the ground truth image, $\mathcal{L}_{\text{MSE}}$ is the normalized pixel-wise L2 loss, $\mathcal{L}_{\text{LPIPS}}$ is perceptual image patch similarity \cite{zhang_lpips_2018}, and $\alpha$ is a loss weighting coefficient.

\section{Experiments}
\label{sec:experiments}

In this section, we outline the experimental setup, followed by an evaluation of our implementation for two-view reconstruction on two synthetic object datasets.
We then highlight the models performance on extrapolated views, and ablate our model components.

\subsection{Experimental Setup}
\label{sec:experimental setup}

\paragraph{Compared Methods.} We compare the following methods: (1) Ours w/o bias, which is our model without the distance attention bias in the transformer, using only DINOv2 features and \plucker coordinate positional encodings with two input views; (2) OpenLRM \cite{openlrm}, an open-source implementation of Large Reconstruction Models (LRM) \cite{hong_lrm_2023} that infers a triplane representation from a single posed image, trained on a single object category; (3) PixelNeRF \cite{yu_pixelnerf_2021}, an extension of NeRF for sparse input views that conditions on image features; and (4) Splatter Image \cite{szymanowicz_splatter_2024}, a state-of-the-art method which uses Gaussian splatting for single-view and few-view category-level reconstruction.

\paragraph{Datasets.}
We evaluate the methods on the Shapenet-SRN `Cars' and `Chairs' datasets \cite{sitzmann2019scene}.
Shapenet-SRN is a synthetic dataset of RGB images taken from a fixed distance around a computer model of an object, which includes camera pose information. 
The datasets have a predefined train/val/test split, and additional details about the datasets is included in the supplement.
We use the images, camera intrinsics, camera poses and data splits as provided by the datasets and train our method using relative camera poses.

\paragraph{Metrics.} We compare novel views with a held-out test set of ground-truth views. 
We measure the perceptual quality (LPIPS) \cite{zhang_lpips_2018}, structural similarity (SSIM), and peak signal-to-noise ratio (PSNR) of these renders. 
For evaluation, we follow the protocol in other similar works using the Shapenet-SRN dataset \cite{yu_pixelnerf_2021, szymanowicz_splatter_2024}, by using view 64 and 128 as the input views.
All unseen views are used for the computation of the metrics, excluding the conditioning input views. 

\paragraph{Implementation and Training Details.}
We train an 8 layer transformer decoder model with a hidden dimension of 512.
We instantiate the linear projection of our input tokens as the identity matrix, with an initial bias of zero.
Each cross-attention and self-attention layer has a learnable parameter ($\gamma$) for scaling the distance bias.

Our image encoder is initialized as a pre-trained small DINOv2 vision transformer \cite{oquab_dinov2_2023}.
We train with $n=2$ random input views of our instance, and $k=2$ ground-truth comparison views (see \cref{sec:training-objective}), and compute camera poses relative to the first.
This is also done at inference time.
Additional training details are available in the supplement, and code is available on the project page \footnote{\href{https://github.com/SamBahrami/PluckeRF}{https://github.com/SamBahrami/PluckeRF}}.

\subsection{Results}

\paragraph{Quantitative Evaluation.}
We report our quantitative results for Shapenet-SRN `Chairs' and `Cars' in \cref{tab:shapenet-experiments}.
Our model outperforms the baseline without the distance bias (Ours w/o bias), demonstrating that incorporating the distance bias enhances the novel view synthesis quality.
Our method outperforms the 2-view implementations of Splatter Image and pixelNeRF for `Chairs', and is similar in performance for the `Cars' dataset.

\paragraph{Qualitative Results.} We provide a qualitative comparison of the compared methods for the `Chairs' dataset in \cref{fig:shapenet-chairs-qualitative} and `Cars' dataset in \cref{fig:shapenet-cars-qualitative}, and additional qualitative results can be seen in the supplement.
We can see that while our method has some blur around the unseen parts of the object, it is notably much crisper than the other methods.
This is a limitation of feed-forward, diffusion-free methods in general.

Our model is challenged by areas where there is a sharp color change, such as on stripes of a car. 
We also note that it struggles with areas that have very fine detail such as detailed textures and geometry.
These limitations may come from how the images are processed, by first converting them to lower resolution patch-wise feature maps, which may lose detail information in the process. 
The orthogonal feature grid is also a limiting factor since it is constrained to its grid resolution, making it challenging to capture fine details without increasing the grid size, which would have an additional memory cost.

\input{qualitative_chairs_figure}
\input{qualitative_cars_figure}

\begin{table}[!t]
\footnotesize
    \caption{
    Novel view synthesis results on the Chairs and Cars Shapenet-SRN test sets.
    Our model is given $n$ input views, and the output renders are evaluated against all other images in the test split.
    Unreported results are marked with a dash.
    Best results are denoted in bold.
    }
    \label{tab:shapenet-experiments}
    \centering
    \setlength{\tabcolsep}{2pt}
    \begin{tabularx}{\linewidth}{@{}lcCCCCCC@{}}
        \toprule
        &&
        \multicolumn{3}{c}{\textbf{SRN Chairs}} & \multicolumn{3}{c}{\textbf{SRN Cars}} \\
        Method & $n$ & PSNR$\uparrow$ & LPIPS$\downarrow$ & SSIM$\uparrow$ & PSNR$\uparrow$ & LPIPS$\downarrow$ & SSIM$\uparrow$ \\
        \midrule
        OpenLRM \cite{openlrm} & 1 & 24.44 & 0.076 & 0.93 & 22.73 & 0.096 & 0.90 \\
        pixelNeRF \cite{yu_pixelnerf_2021} & 1 & 23.72 & 0.128 & 0.90 & 23.17 & 0.146 & 0.89 \\
        SplatterImg \cite{szymanowicz_splatter_2024} & 1 & 24.43 & 0.067 & 0.93 & 24.00 & 0.078 & 0.92 \\
        \midrule
        pixelNeRF \cite{yu_pixelnerf_2021} & 2 & 25.97 & 0.071 & 0.94 & 25.66 & 0.079 & \textbf{0.94} \\
        SplatterImg \cite{szymanowicz_splatter_2024} & 2 & 25.72 & 0.056 & 0.94 & \textbf{26.01} & -- & \textbf{0.94} \\
        \midrule
        Ours w/o bias & 2 & 27.67 & 0.048 & \textbf{0.96} & 25.26 & 0.073 & 0.93 \\
        Ours & 2 & \textbf{28.22} & \textbf{0.045} & \textbf{0.96} & 25.54 & \textbf{0.070} & \textbf{0.94} \\
        \bottomrule
    \end{tabularx}
\end{table}

\subsection{Evaluating Extrapolated Views}

In our qualitative results (\cref{fig:shapenet-chairs-qualitative} and \cref{fig:shapenet-cars-qualitative}), our novel views visually contain more high-frequency detail than the compared methods (pixelNeRF and Splatter Image), especially on the unseen side of the car.
To quantify this, we re-evaluate model performance on the subset of extrapolated novel views, which we define as any view with an out-of-plane rotation that is at least $90^{\circ}$ from both input views.
We calculate this by computing the angle between the rotations of each input view with those of the other views in the test set, noting that the test set does not include any in-plane rotations (\ie, rotations about the camera optical axis).
The results on these extrapolated views can be seen in \cref{tab:extrapolation-results}.
Our method outperforms the other compared feed-forward methods more significantly under these conditions.

\begin{table}[!t]
    \small
    \caption{
    Extrapolation performance.
    We evaluate on a subset of the Shapenet-SRN Chairs and Cars test set, corresponding to novel views that are rotated out-of-plane by at least $90^{\circ}$ from both input views.
    That is, they are likely to observe the \textit{unseen} side of the object, rather than views that interpolate the input set.
    We use the standard pair of input views used in the main experiments, and one view in the OpenLRM case.
    Bold indicates best result, unreported results are marked with a dash.
    }
    \label{tab:extrapolation-results}
    \centering
    \setlength{\tabcolsep}{1pt}
    \begin{tabularx}{\linewidth}{@{}lCCCCCC@{}}
        \toprule
        & \multicolumn{3}{c}{\textbf{SRN Chairs}} & \multicolumn{3}{c}{\textbf{SRN Cars}} \\
        Method & PSNR$\uparrow$ & LPIPS$\downarrow$ & SSIM$\uparrow$ & PSNR$\uparrow$ & LPIPS$\downarrow$ & SSIM$\uparrow$ \\
        \midrule
        OpenLRM \cite{openlrm} & 24.24 & 0.080 & 0.926 & 22.02 & 0.104 & 0.893 \\
        \midrule
        pixelNeRF \cite{yu_pixelnerf_2021} & 25.33 & 0.079 & 0.939 & 23.65 & 0.105 & 0.916 \\
        SplatterImg \cite{szymanowicz_splatter_2024} & 25.15 & 0.064 & 0.936 & -- & -- & -- \\
        \midrule
        Ours w/o bias & 27.48 & 0.050 & 0.956 & 24.68 & 0.080 & 0.926\\
        Ours & \textbf{27.97} & \textbf{0.047} & \textbf{0.959} & \textbf{24.97} & \textbf{0.076} & \textbf{0.928} \\
        \bottomrule
    \end{tabularx}
\end{table}

\subsection{Ablation Study and Analysis}
\label{sec:ablations}
We conducted a series of ablation experiments to evaluate the influence of individual components of our method on the final performance. 
Due to computational cost, we train these models at a shorter training schedule for 60k iterations with a value of $\alpha$ set to 0 initially, and 0.01 for the final 20k steps on the Shapenet-SRN `Chairs' dataset.
Additionally we set our orthogonal feature grid resolution $M$ to 48 instead of 64 which was used in the main experiments.
All other training parameters remained consistent between these ablations and the primary experiments.
We show the results of our ablation study for the two-view model in \cref{tab:ablation-components}.

We ablate the design choice of finetuning the image encoder by instead freezing these parameters (``w/o DINOv2 finetuning'') and observe a significant decrease in the model's performance.
We ablate the choice of setting the input tokens of the transformer to a linear projection of our \method lines by instead using learnable tokens of the same size (``w/o \method input'').
While this has a negligible impact on the final performance of the model, in our experiments the model converges in fewer steps with \method initialization.
We ablate the choice of concatenating the \plucker coordinates to the image patch features by instead using image features only (``w/o \plucker encoding''), and observe a small drop in the model's novel view synthesis performance.
We hypothesize that the distance bias already provides most of the positional information needed by the model. 
We ablate the use of the perceptual similarity loss (``w/o $\mathcal{L}_{\text{LPIPS}}$'') and see that the resulting model has a similar PSNR, but also blurrier images and therefore a notable deterioration of the perceptual similarity.
Finally, we ablate the learnable scaling parameters $\gamma$ by setting them to 1 (``w/o learnable $\gamma$'') and observe a significant drop in performance.

\begin{table}[!t]
    \small
  \setlength{\tabcolsep}{2pt}
  \caption{Ablation study.
  Experiments were undertaken on the Shapenet-SRN Chairs dataset, with two input views. 
  The ablations are conducted on a shorter training run than the main model results, with a smaller orthogonal feature grid representation, and evaluated on the same test data split as the main experiments. 
  Bold values indicate the highest performance in that metric, underline indicates second best.}
  \label{tab:ablation-components}
  \centering
    \begin{tabularx}{\linewidth}{@{}lCCC@{}}
    \toprule
    Method & PSNR$\uparrow$ & LPIPS$\downarrow$ & SSIM$\uparrow$ \\
    \midrule
    Ours & \textbf{24.21} & \textbf{0.091} & \textbf{0.926} \\
    w/o \method input & \underline{24.17} & 0.093 & \underline{0.925} \\
    w/o \plucker positional encoding & 23.96 & \textbf{0.091} & 0.924 \\
    w/o $\mathcal{L}_{\text{LPIPS}}$ & 24.10 & 0.102 & 0.922 \\
    w/o distance bias, \cref{eq:distance-weighting} & 23.44 & 0.100 & 0.917 \\
    w/o DINOv2 finetuning & 23.10 & 0.101 & 0.913 \\
    w/o learnable $\gamma$, \cref{eq:distance-weighting} & 22.83 & 0.109 & 0.909 \\
    \bottomrule
  \end{tabularx}
\end{table}

\section{Discussion and Conclusion}
\label{sec:discussion-conclusion}

We observe that adding our 2D--3D and 3D--3D distance biases into the attention blocks of a feed-forward few-view reconstruction transformer significantly improves its performance. 
However, this spatial locality bias may make it more challenging for our model to leverage symmetries in the dataset, since these would allow information to be shared non-locally.
In addition, our approach is NeRF-based, which is slower to render and more memory intensive compared to other representations, such as Gaussian mixtures \cite{kerbl_3d_2023, smart2024splatt3r, zhang2024gslrmlargereconstructionmodel}.
More fundamentally, feed-forward direct prediction methods are prone to blurriness in unseen regions, since they average over the possible completions, unlike diffusion-based methods that are slower but sharp.

In this paper, we introduced \method, a method for few-view 3D reconstruction that integrates 2D image information into a 3D orthogonal feature grid through a line--line distance-based attention bias.
This distance bias operates (1) between input image rays and the 3D \method representation in cross-attention layers, and (2) within the 3D representation in self-attention layers, and acts to create an inductive bias in the model to help it solve the data-to-model association problem.
Our model can rapidly infer a plausible 3D representation from only a few input views, outperforming existing approaches.
Future work may consider a more compute-efficient 3D representation like a Gaussian mixture or introduce these inductive biases into a diffusion transformer for blur-free reconstruction.

\newpage
{
    \small
    \bibliographystyle{ieeenat_fullname}
    \bibliography{main}
}

\clearpage
\setcounter{page}{1}
\maketitlesupplementary
\appendix %

In this appendix we present additional qualitative results, training details for compared methods, and details about the training and inference of our trained models.

\section{Additional Results}
\input{supplement_chairs_results}
\input{supplement_results}
We present additional qualitative static comparisons on ShapeNet-SRN `Chairs' and `Cars' including more views than shown in the main results.
We compare our method to OpenLRM \cite{openlrm} which has been trained on a single object category, pixelNeRF \cite{yu_pixelnerf_2021}, Splatter Image \cite{szymanowicz_splatter_2024}, our baseline without bias, and our main method in \cref{fig:supplement_chairs_results} and \cref{fig:supplement_results}.

\section{Shapenet SRN Dataset Details}
The Shapenet SRN `Cars' dataset contains 3514 instances, with a predefined train/val/test split of 2458/352/704 instances respectively.
The `Chairs' dataset contains 6519 instances with a 4612/662/1317 train/val/test split.
Each training instance contains 50 posed images taken from random points on a sphere around the object.
Each testing and validation instance contains 251 posed images taken on an Archimedean spiral along the sphere.
All scenes share camera intrinsics, and images are rendered at a resolution of (128, 128).

\section{Model Training and Inference Details}
We use the AdamW \cite{loshchilov_decoupled_2019} optimizer with weight decay of 0.05 and \( \beta_1 \) of 0.9, \( \beta_2 \) of 0.95. 
Our models were trained on 4 GPUs (Nvidia A100-40GB) with a batch size of 32
and with separate models for each object category.
Training takes 5 days on 4 A100-40GB GPUs for 500k training steps. 
Inference takes roughly 15 seconds to render all 251 novel test views for an image pair on a single A100. 

Our image encoder is initialized as a pre-trained small DINOv2 vision transformer \cite{oquab_dinov2_2023}, which has a token size of 384 (\texttt{dinov2\_vits14\_reg}) and creates image patches of size $14 \times 14$ pixels.
We rescale our input images to $448 \times 448$ pixels using bicubic interpolation to get $1024$ patches per image.

We use a maximum learning rate of \(4 \times 10^{-4}\) for our `Chairs' model, and \(8 \times 10^{-5}\) for the `Cars' model, and
a cosine scheduler for the learning rate, with 2500 warm-up iterations. 
We train our `Cars' model for 500k iterations, and set the LPIPS loss weight $\alpha$ to 0 at the start of training and increase it to 0.01 at iteration 400k.
We train our `Chairs' model for 800k iterations, and set $\alpha$ to 0.01 at iteration 650k.

\section{Splatter Image Training Details}
We use the official Splatter Image implementation \cite{szymanowicz_splatter_2024}.
Our 2 view Splatter Image model for the Shapenet-SRN `Chairs' category was trained with their 2 view training configuration recommendations with a L40S GPU for 800k iterations with a batch size of 8, and then fine-tuned with an LPIPS loss for 100k additional training iterations. 

\section{OpenLRM Training Details}
We trained our single category OpenLRM models with the same training parameters as the main experiment where possible.
We freeze the image encoder as per the original LRM architecture \cite{hong_lrm_2023} and use the same image encoder model (\texttt{dinov2\_vits14\_reg}) as our main experiments.
We used the same training loss as our Shapenet-SRN `Cars' model, setting $\alpha$ to 0.01 at training step 400k, and trained each model for a total of 500k steps.
From our experiments, setting a greater emphasis on the $\mathcal{L}_{\text{MSE}}$ loss resulted in higher PSNR in the final novel views.

\end{document}

%% file: preamble.tex
\newcommand{\method}{Pl\"uckeRF\xspace} %
\newcommand{\plucker}{Pl\"ucker\xspace}

\newif\ifincludeappendix

%% file: macros.tex
\usepackage{xspace}
\makeatletter
\DeclareRobustCommand\onedot{\futurelet\@let@token\@onedot}
\def\@onedot{\ifx\@let@token.\else.\null\fi\xspace}

\def\eg{e.g\onedot} 
\def\ie{i.e\onedot}

\def\etal{et al\onedot}
\makeatother

\newcommand{\reals}{\mathbb{R}}

\newcommand{\transpose}{^\mathsf{T}}

\DeclareMathOperator{\softmax}{softmax}

\newcommand{\bd}{\mathbf{d}}

\newcommand{\bbf}{\mathbf{f}}

\newcommand{\bl}{{\mathbf{l}}}
\newcommand{\bem}{{\mathbf{m}}}

\newcommand{\bo}{{\mathbf{o}}}
\newcommand{\bp}{{\mathbf{p}}}

\newcommand{\bt}{{\mathbf{t}}}

\newcommand{\bD}{{\mathbf{D}}}

\newcommand{\bK}{{\mathbf{K}}}

\newcommand{\bQ}{{\mathbf{Q}}}
\newcommand{\bR}{{\mathbf{R}}}

\newcommand{\bT}{{\mathbf{T}}}

\newcommand{\bV}{{\mathbf{V}}}

%% file: qualitative_chairs_figure.tex
\begin{figure}[!t]
  \centering
\centering
    \makebox[\linewidth]{
        \begin{subfigure}{0.15\linewidth}
            \centering
            \footnotesize Input View Pair
        \end{subfigure}
        \begin{subfigure}{0.15\linewidth}
            \centering
            \footnotesize pixelNeRF
        \end{subfigure}
        \begin{subfigure}{0.15\linewidth}
            \centering
            \footnotesize SplatterImg
        \end{subfigure}
        \begin{subfigure}{0.15\linewidth}
            \centering
            \footnotesize Ours w/o bias
        \end{subfigure}
        \begin{subfigure}{0.15\linewidth}
            \centering
            \footnotesize Ours
        \end{subfigure}
        \begin{subfigure}{0.15\linewidth}
            \centering
            \footnotesize GT
        \end{subfigure}
    }

    \begin{subfigure}{0.15\linewidth}
        \centering
        \includegraphics[width=\textwidth, trim=15pt 15pt 15pt 15pt, clip]{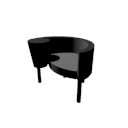}
    \end{subfigure}
    \begin{subfigure}{0.15\linewidth}
        \centering
        \includegraphics[width=\textwidth, trim=15pt 15pt 15pt 15pt, clip]{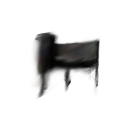}
    \end{subfigure}
    \begin{subfigure}{0.15\linewidth}
        \centering
        \includegraphics[width=\textwidth, trim=15pt 15pt 15pt 15pt, clip]{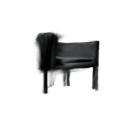}
    \end{subfigure}
    \begin{subfigure}{0.15\linewidth}
        \centering
        \includegraphics[width=\textwidth, trim=15pt 15pt 15pt 15pt, clip]{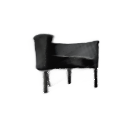}
    \end{subfigure}
    \begin{subfigure}{0.15\linewidth}
        \centering
        \includegraphics[width=\textwidth, trim=15pt 15pt 15pt 15pt, clip]{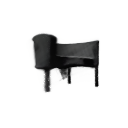}
    \end{subfigure}
    \begin{subfigure}{0.15\linewidth}
        \centering
        \includegraphics[width=\textwidth, trim=15pt 15pt 15pt 15pt, clip]{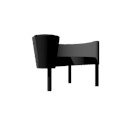}
    \end{subfigure}

    \begin{subfigure}{0.15\linewidth}
        \centering
        \includegraphics[width=\textwidth, trim=15pt 15pt 15pt 15pt, clip]{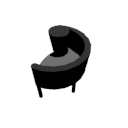}
    \end{subfigure}
    \begin{subfigure}{0.15\linewidth}
        \centering
        \includegraphics[width=\textwidth, trim=15pt 15pt 15pt 15pt, clip]{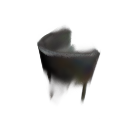}
    \end{subfigure}
    \begin{subfigure}{0.15\linewidth}
        \centering
        \includegraphics[width=\textwidth, trim=15pt 15pt 15pt 15pt, clip]{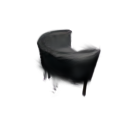}
    \end{subfigure}
    \begin{subfigure}{0.15\linewidth}
        \centering
        \includegraphics[width=\textwidth, trim=15pt 15pt 15pt 15pt, clip]{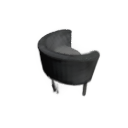}
    \end{subfigure}
    \begin{subfigure}{0.15\linewidth}
        \centering
        \includegraphics[width=\textwidth, trim=15pt 15pt 15pt 15pt, clip]{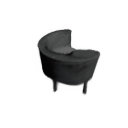}
    \end{subfigure}
    \begin{subfigure}{0.15\linewidth}
        \centering
        \includegraphics[width=\textwidth, trim=15pt 15pt 15pt 15pt, clip]{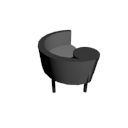}
    \end{subfigure}

        \begin{subfigure}{0.15\linewidth}
        \centering
        \includegraphics[width=\textwidth, trim=15pt 15pt 15pt 15pt, clip]{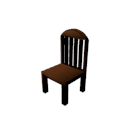}
    \end{subfigure}
    \begin{subfigure}{0.15\linewidth}
        \centering
        \includegraphics[width=\textwidth, trim=15pt 15pt 15pt 15pt, clip]{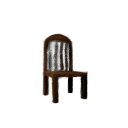}
    \end{subfigure}
    \begin{subfigure}{0.15\linewidth}
        \centering
        \includegraphics[width=\textwidth, trim=15pt 15pt 15pt 15pt, clip]{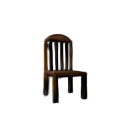}
    \end{subfigure}
    \begin{subfigure}{0.15\linewidth}
        \centering
        \includegraphics[width=\textwidth, trim=15pt 15pt 15pt 15pt, clip]{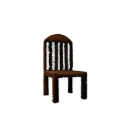}
    \end{subfigure}
    \begin{subfigure}{0.15\linewidth}
        \centering
        \includegraphics[width=\textwidth, trim=15pt 15pt 15pt 15pt, clip]{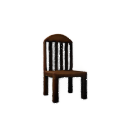}
    \end{subfigure}
    \begin{subfigure}{0.15\linewidth}
        \centering
        \includegraphics[width=\textwidth, trim=15pt 15pt 15pt 15pt, clip]{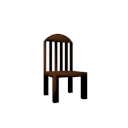}
    \end{subfigure}

    \begin{subfigure}{0.15\linewidth}
        \centering
        \includegraphics[width=\textwidth, trim=15pt 15pt 15pt 15pt, clip]{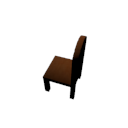}
    \end{subfigure}
    \begin{subfigure}{0.15\linewidth}
        \centering
        \includegraphics[width=\textwidth, trim=15pt 15pt 15pt 15pt, clip]{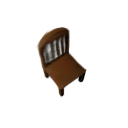}
    \end{subfigure}
    \begin{subfigure}{0.15\linewidth}
        \centering
        \includegraphics[width=\textwidth, trim=15pt 15pt 15pt 15pt, clip]{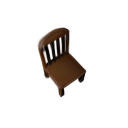}
    \end{subfigure}
    \begin{subfigure}{0.15\linewidth}
        \centering
        \includegraphics[width=\textwidth, trim=15pt 15pt 15pt 15pt, clip]{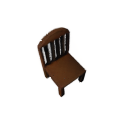}
    \end{subfigure}
    \begin{subfigure}{0.15\linewidth}
        \centering
        \includegraphics[width=\textwidth, trim=15pt 15pt 15pt 15pt, clip]{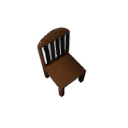}
    \end{subfigure}
    \begin{subfigure}{0.15\linewidth}
        \centering
        \includegraphics[width=\textwidth, trim=15pt 15pt 15pt 15pt, clip]{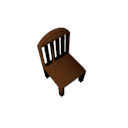}
    \end{subfigure}

        \begin{subfigure}{0.15\linewidth}
        \centering
        \includegraphics[width=\textwidth, trim=15pt 15pt 15pt 15pt, clip]{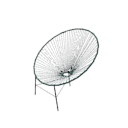}
    \end{subfigure}
    \begin{subfigure}{0.15\linewidth}
        \centering
        \includegraphics[width=\textwidth, trim=15pt 15pt 15pt 15pt, clip]{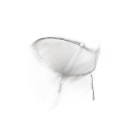}
    \end{subfigure}
    \begin{subfigure}{0.15\linewidth}
        \centering
        \includegraphics[width=\textwidth, trim=15pt 15pt 15pt 15pt, clip]{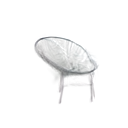}
    \end{subfigure}
    \begin{subfigure}{0.15\linewidth}
        \centering
        \includegraphics[width=\textwidth, trim=15pt 15pt 15pt 15pt, clip]{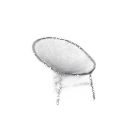}
    \end{subfigure}
    \begin{subfigure}{0.15\linewidth}
        \centering
        \includegraphics[width=\textwidth, trim=15pt 15pt 15pt 15pt, clip]{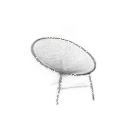}
    \end{subfigure}
    \begin{subfigure}{0.15\linewidth}
        \centering
        \includegraphics[width=\textwidth, trim=15pt 15pt 15pt 15pt, clip]{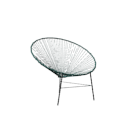}
    \end{subfigure}

    \begin{subfigure}{0.15\linewidth}
        \centering
        \includegraphics[width=\textwidth, trim=15pt 15pt 15pt 15pt, clip]{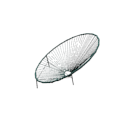}
    \end{subfigure}
    \begin{subfigure}{0.15\linewidth}
        \centering
        \includegraphics[width=\textwidth, trim=15pt 15pt 15pt 15pt, clip]{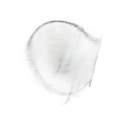}
    \end{subfigure}
    \begin{subfigure}{0.15\linewidth}
        \centering
        \includegraphics[width=\textwidth, trim=15pt 15pt 15pt 15pt, clip]{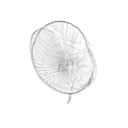}
    \end{subfigure}
    \begin{subfigure}{0.15\linewidth}
        \centering
        \includegraphics[width=\textwidth, trim=15pt 15pt 15pt 15pt, clip]{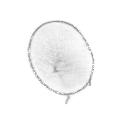}
    \end{subfigure}
    \begin{subfigure}{0.15\linewidth}
        \centering
        \includegraphics[width=\textwidth, trim=15pt 15pt 15pt 15pt, clip]{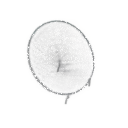}
    \end{subfigure}
    \begin{subfigure}{0.15\linewidth}
        \centering
        \includegraphics[width=\textwidth, trim=15pt 15pt 15pt 15pt, clip]{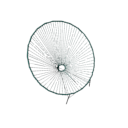}
    \end{subfigure}

    \caption{Shapenet-SRN Chairs qualitative comparison. The model is given two fixed input views, and renders novel views corresponding to the other ground truth views in the test set around the object. Our method is able to better infer the shape and textures of the unseen sides of the chairs, and produce sharper outputs. Please zoom in to observe finer details.}
    \label{fig:shapenet-chairs-qualitative}
\end{figure}

%% file: qualitative_cars_figure.tex
\begin{figure}[!t]
  \centering
\centering
    \makebox[\linewidth]{
        \begin{subfigure}{0.15\linewidth}
            \centering
            \footnotesize Input View Pair
        \end{subfigure}
        \begin{subfigure}{0.15\linewidth}
            \centering
            \footnotesize pixelNeRF
        \end{subfigure}
        \begin{subfigure}{0.15\linewidth}
            \centering
            \footnotesize SplatterImg ($n=1$)
        \end{subfigure}
        \begin{subfigure}{0.15\linewidth}
            \centering
            \footnotesize Ours w/o bias
        \end{subfigure}
        \begin{subfigure}{0.15\linewidth}
            \centering
            \footnotesize Ours
        \end{subfigure}
        \begin{subfigure}{0.15\linewidth}
            \centering
            \footnotesize GT
        \end{subfigure}
    }

    \begin{subfigure}{0.15\linewidth}
        \centering
        \includegraphics[width=\textwidth, trim=15pt 15pt 15pt 15pt, clip]{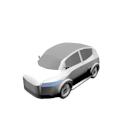}
    \end{subfigure}
    \begin{subfigure}{0.15\linewidth}
        \centering
        \includegraphics[width=\textwidth, trim=15pt 15pt 15pt 15pt, clip]{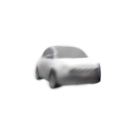}
    \end{subfigure}
    \begin{subfigure}{0.15\linewidth}
        \centering
        \includegraphics[width=\textwidth, trim=15pt 15pt 15pt 15pt, clip]{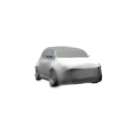}
    \end{subfigure}
    \begin{subfigure}{0.15\linewidth}
        \centering
        \includegraphics[width=\textwidth, trim=15pt 15pt 15pt 15pt, clip]{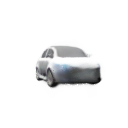}
    \end{subfigure}
    \begin{subfigure}{0.15\linewidth}
        \centering
        \includegraphics[width=\textwidth, trim=15pt 15pt 15pt 15pt, clip]{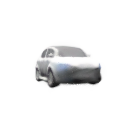}
    \end{subfigure}
    \begin{subfigure}{0.15\linewidth}
        \centering
        \includegraphics[width=\textwidth, trim=15pt 15pt 15pt 15pt, clip]{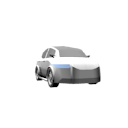}
    \end{subfigure}

    \begin{subfigure}{0.15\linewidth}
        \centering
        \includegraphics[width=\textwidth, trim=15pt 15pt 15pt 15pt, clip]{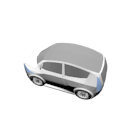}
    \end{subfigure}
    \begin{subfigure}{0.15\linewidth}
        \centering
        \includegraphics[width=\textwidth, trim=15pt 15pt 15pt 15pt, clip]{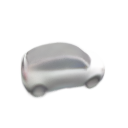}
    \end{subfigure}
    \begin{subfigure}{0.15\linewidth}
        \centering
        \includegraphics[width=\textwidth, trim=15pt 15pt 15pt 15pt, clip]{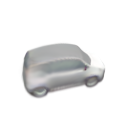}
    \end{subfigure}
    \begin{subfigure}{0.15\linewidth}
        \centering
        \includegraphics[width=\textwidth, trim=15pt 15pt 15pt 15pt, clip]{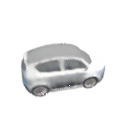}
    \end{subfigure}
    \begin{subfigure}{0.15\linewidth}
        \centering
        \includegraphics[width=\textwidth, trim=15pt 15pt 15pt 15pt, clip]{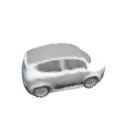}
    \end{subfigure}
    \begin{subfigure}{0.15\linewidth}
        \centering
        \includegraphics[width=\textwidth, trim=15pt 15pt 15pt 15pt, clip]{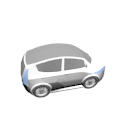}
    \end{subfigure}

    \begin{subfigure}{0.15\linewidth}
        \centering
        \includegraphics[width=\textwidth, trim=15pt 15pt 15pt 15pt, clip]{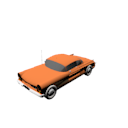}
    \end{subfigure}
    \begin{subfigure}{0.15\linewidth}
        \centering
        \includegraphics[width=\textwidth, trim=15pt 15pt 15pt 15pt, clip]{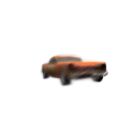}
    \end{subfigure}
    \begin{subfigure}{0.15\linewidth}
        \centering
        \includegraphics[width=\textwidth, trim=15pt 15pt 15pt 15pt, clip]{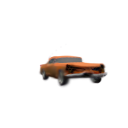}
    \end{subfigure}
    \begin{subfigure}{0.15\linewidth}
        \centering
        \includegraphics[width=\textwidth, trim=15pt 15pt 15pt 15pt, clip]{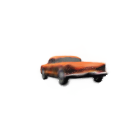}
    \end{subfigure}
    \begin{subfigure}{0.15\linewidth}
        \centering
        \includegraphics[width=\textwidth, trim=15pt 15pt 15pt 15pt, clip]{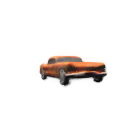}
    \end{subfigure}
    \begin{subfigure}{0.15\linewidth}
        \centering
        \includegraphics[width=\textwidth, trim=15pt 15pt 15pt 15pt, clip]{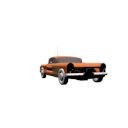}
    \end{subfigure}

    \begin{subfigure}{0.15\linewidth}
        \centering
        \includegraphics[width=\textwidth, trim=15pt 15pt 15pt 15pt, clip]{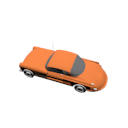}
    \end{subfigure}
    \begin{subfigure}{0.15\linewidth}
        \centering
        \includegraphics[width=\textwidth, trim=15pt 15pt 15pt 15pt, clip]{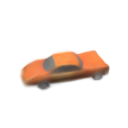}
    \end{subfigure}
    \begin{subfigure}{0.15\linewidth}
        \centering
        \includegraphics[width=\textwidth, trim=15pt 15pt 15pt 15pt, clip]{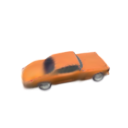}
    \end{subfigure}
    \begin{subfigure}{0.15\linewidth}
        \centering
        \includegraphics[width=\textwidth, trim=15pt 15pt 15pt 15pt, clip]{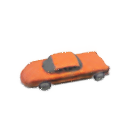}
    \end{subfigure}
    \begin{subfigure}{0.15\linewidth}
        \centering
        \includegraphics[width=\textwidth, trim=15pt 15pt 15pt 15pt, clip]{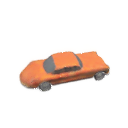}
    \end{subfigure}
    \begin{subfigure}{0.15\linewidth}
        \centering
        \includegraphics[width=\textwidth, trim=15pt 15pt 15pt 15pt, clip]{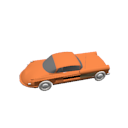}
    \end{subfigure}
    \caption{Shapenet-SRN Cars qualitative comparison. The model is given two fixed input views, and renders novel views corresponding to the other ground truth views in the testing set around the object. Note that the Splatter Image images in this comparison are from their single view model ($n = 1$). Please zoom in to observe finer details.}
    \label{fig:shapenet-cars-qualitative}
\end{figure}

%% file: supplement_chairs_results.tex
\begin{figure*}[!t]
    \centering
    \begin{minipage}{0.75\linewidth}\centering
        \begin{subfigure}{0.13\linewidth}
            \centering
            Input View Pair
        \end{subfigure}
        \begin{subfigure}{0.13\linewidth}
            \centering
            OpenLRM
        \end{subfigure}
        \begin{subfigure}{0.13\linewidth}
            \centering
            pixelNeRF
        \end{subfigure}
        \begin{subfigure}{0.13\linewidth}
            \centering
            Splatter Image
        \end{subfigure}
        \begin{subfigure}{0.13\linewidth}
            \centering
            Ours w/o bias
        \end{subfigure}
        \begin{subfigure}{0.13\linewidth}
            \centering
            Ours
        \end{subfigure}
        \begin{subfigure}{0.13\linewidth}
            \centering
            Ground Truth
        \end{subfigure}\vfill    

        \begin{subfigure}{0.13\linewidth}
            \centering
            \includegraphics[width=\textwidth, trim=15pt 15pt 15pt 15pt, clip]{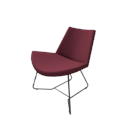}
        \end{subfigure}
        \begin{subfigure}{0.13\linewidth}
            \centering
            \includegraphics[width=\textwidth, trim=15pt 15pt 15pt 15pt, clip]{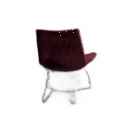}
        \end{subfigure}
        \begin{subfigure}{0.13\linewidth}
            \centering
            \includegraphics[width=\textwidth, trim=15pt 15pt 15pt 15pt, clip]{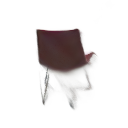}
        \end{subfigure}
        \begin{subfigure}{0.13\linewidth}
            \centering
            \includegraphics[width=\textwidth, trim=15pt 15pt 15pt 15pt, clip]{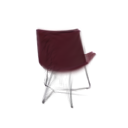}
        \end{subfigure}
        \begin{subfigure}{0.13\linewidth}
            \centering
            \includegraphics[width=\textwidth, trim=15pt 15pt 15pt 15pt, clip]{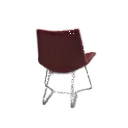}
        \end{subfigure}
        \begin{subfigure}{0.13\linewidth}
            \centering
            \includegraphics[width=\textwidth, trim=15pt 15pt 15pt 15pt, clip]{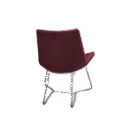}
        \end{subfigure}
        \begin{subfigure}{0.13\linewidth}
            \centering
            \includegraphics[width=\textwidth, trim=15pt 15pt 15pt 15pt, clip]{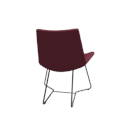}
        \end{subfigure}
    
        \vspace{-4pt}
    
        \begin{subfigure}{0.13\linewidth}
            \centering
            \includegraphics[width=\textwidth, trim=15pt 15pt 15pt 15pt, clip]{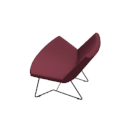}
        \end{subfigure}
        \begin{subfigure}{0.13\linewidth}
            \centering
            \includegraphics[width=\textwidth, trim=15pt 15pt 15pt 15pt, clip]{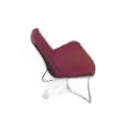}
        \end{subfigure}
        \begin{subfigure}{0.13\linewidth}
            \centering
            \includegraphics[width=\textwidth, trim=15pt 15pt 15pt 15pt, clip]{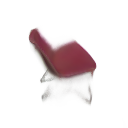}
        \end{subfigure}
        \begin{subfigure}{0.13\linewidth}
            \centering
            \includegraphics[width=\textwidth, trim=15pt 15pt 15pt 15pt, clip]{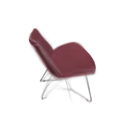}
        \end{subfigure}
        \begin{subfigure}{0.13\linewidth}
            \centering
            \includegraphics[width=\textwidth, trim=15pt 15pt 15pt 15pt, clip]{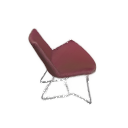}
        \end{subfigure}
        \begin{subfigure}{0.13\linewidth}
            \centering
            \includegraphics[width=\textwidth, trim=15pt 15pt 15pt 15pt, clip]{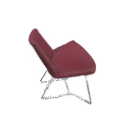}
        \end{subfigure}
        \begin{subfigure}{0.13\linewidth}
            \centering
            \includegraphics[width=\textwidth, trim=15pt 15pt 15pt 15pt, clip]{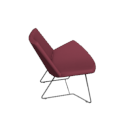}
        \end{subfigure}
    
        \vspace{-4pt}
    
        \begin{subfigure}{0.13\linewidth}
            \centering
            \includegraphics[width=\textwidth, trim=15pt 15pt 15pt 15pt, clip]{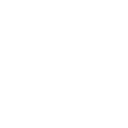}
        \end{subfigure}
        \begin{subfigure}{0.13\linewidth}
            \centering
            \includegraphics[width=\textwidth, trim=15pt 15pt 15pt 15pt, clip]{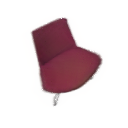}
        \end{subfigure}
        \begin{subfigure}{0.13\linewidth}
            \centering
            \includegraphics[width=\textwidth, trim=15pt 15pt 15pt 15pt, clip]{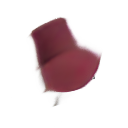}
        \end{subfigure}
        \begin{subfigure}{0.13\linewidth}
            \centering
            \includegraphics[width=\textwidth, trim=15pt 15pt 15pt 15pt, clip]{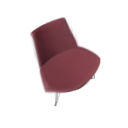}
        \end{subfigure}
        \begin{subfigure}{0.13\linewidth}
            \centering
            \includegraphics[width=\textwidth, trim=15pt 15pt 15pt 15pt, clip]{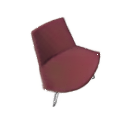}
        \end{subfigure}
        \begin{subfigure}{0.13\linewidth}
            \centering
            \includegraphics[width=\textwidth, trim=15pt 15pt 15pt 15pt, clip]{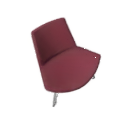}
        \end{subfigure}
        \begin{subfigure}{0.13\linewidth}
            \centering
            \includegraphics[width=\textwidth, trim=15pt 15pt 15pt 15pt, clip]{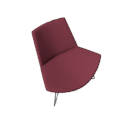}
        \end{subfigure}
    
        \vspace{-4pt}
    
        \begin{subfigure}{0.13\linewidth}
            \centering
            \includegraphics[width=\textwidth, trim=15pt 15pt 15pt 15pt, clip]{render_placeholder.png}
        \end{subfigure}
        \begin{subfigure}{0.13\linewidth}
            \centering
            \includegraphics[width=\textwidth, trim=15pt 15pt 15pt 15pt, clip]{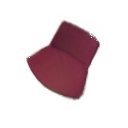}
        \end{subfigure}
        \begin{subfigure}{0.13\linewidth}
            \centering
            \includegraphics[width=\textwidth, trim=15pt 15pt 15pt 15pt, clip]{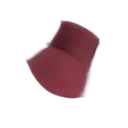}
        \end{subfigure}
        \begin{subfigure}{0.13\linewidth}
            \centering
            \includegraphics[width=\textwidth, trim=15pt 15pt 15pt 15pt, clip]{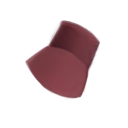}
        \end{subfigure}
        \begin{subfigure}{0.13\linewidth}
            \centering
            \includegraphics[width=\textwidth, trim=15pt 15pt 15pt 15pt, clip]{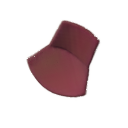}
        \end{subfigure}
        \begin{subfigure}{0.13\linewidth}
            \centering
            \includegraphics[width=\textwidth, trim=15pt 15pt 15pt 15pt, clip]{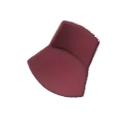}
        \end{subfigure}
        \begin{subfigure}{0.13\linewidth}
            \centering
            \includegraphics[width=\textwidth, trim=15pt 15pt 15pt 15pt, clip]{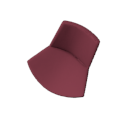}
        \end{subfigure}
    
        \vspace{-2pt}

        \begin{subfigure}{0.13\linewidth}
            \centering
            \includegraphics[width=\textwidth, trim=15pt 15pt 15pt 15pt, clip]{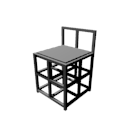}
        \end{subfigure}
        \begin{subfigure}{0.13\linewidth}
            \centering
            \includegraphics[width=\textwidth, trim=15pt 15pt 15pt 15pt, clip]{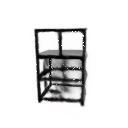}
        \end{subfigure}
        \begin{subfigure}{0.13\linewidth}
            \centering
            \includegraphics[width=\textwidth, trim=15pt 15pt 15pt 15pt, clip]{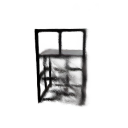}
        \end{subfigure}
        \begin{subfigure}{0.13\linewidth}
            \centering
            \includegraphics[width=\textwidth, trim=15pt 15pt 15pt 15pt, clip]{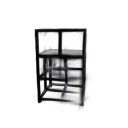}
        \end{subfigure}
        \begin{subfigure}{0.13\linewidth}
            \centering
            \includegraphics[width=\textwidth, trim=15pt 15pt 15pt 15pt, clip]{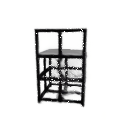}
        \end{subfigure}
        \begin{subfigure}{0.13\linewidth}
            \centering
            \includegraphics[width=\textwidth, trim=15pt 15pt 15pt 15pt, clip]{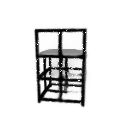}
        \end{subfigure}
        \begin{subfigure}{0.13\linewidth}
            \centering
            \includegraphics[width=\textwidth, trim=15pt 15pt 15pt 15pt, clip]{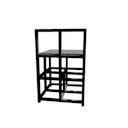}
        \end{subfigure}
    
        \vspace{-4pt}
    
        \begin{subfigure}{0.13\linewidth}
            \centering
            \includegraphics[width=\textwidth, trim=15pt 15pt 15pt 15pt, clip]{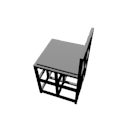}
        \end{subfigure}
        \begin{subfigure}{0.13\linewidth}
            \centering
            \includegraphics[width=\textwidth, trim=15pt 15pt 15pt 15pt, clip]{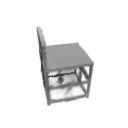}
        \end{subfigure}
        \begin{subfigure}{0.13\linewidth}
            \centering
            \includegraphics[width=\textwidth, trim=15pt 15pt 15pt 15pt, clip]{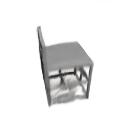}
        \end{subfigure}
        \begin{subfigure}{0.13\linewidth}
            \centering
            \includegraphics[width=\textwidth, trim=15pt 15pt 15pt 15pt, clip]{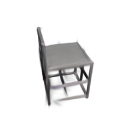}
        \end{subfigure}
        \begin{subfigure}{0.13\linewidth}
            \centering
            \includegraphics[width=\textwidth, trim=15pt 15pt 15pt 15pt, clip]{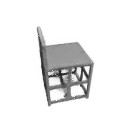}
        \end{subfigure}
        \begin{subfigure}{0.13\linewidth}
            \centering
            \includegraphics[width=\textwidth, trim=15pt 15pt 15pt 15pt, clip]{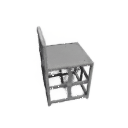}
        \end{subfigure}
        \begin{subfigure}{0.13\linewidth}
            \centering
            \includegraphics[width=\textwidth, trim=15pt 15pt 15pt 15pt, clip]{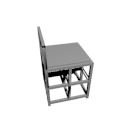}
        \end{subfigure}
    
        \vspace{-4pt}
    
        \begin{subfigure}{0.13\linewidth}
            \centering
            \includegraphics[width=\textwidth, trim=15pt 15pt 15pt 15pt, clip]{render_placeholder.png}
        \end{subfigure}
        \begin{subfigure}{0.13\linewidth}
            \centering
            \includegraphics[width=\textwidth, trim=15pt 15pt 15pt 15pt, clip]{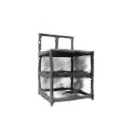}
        \end{subfigure}
        \begin{subfigure}{0.13\linewidth}
            \centering
            \includegraphics[width=\textwidth, trim=15pt 15pt 15pt 15pt, clip]{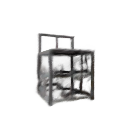}
        \end{subfigure}
        \begin{subfigure}{0.13\linewidth}
            \centering
            \includegraphics[width=\textwidth, trim=15pt 15pt 15pt 15pt, clip]{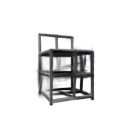}
        \end{subfigure}
        \begin{subfigure}{0.13\linewidth}
            \centering
            \includegraphics[width=\textwidth, trim=15pt 15pt 15pt 15pt, clip]{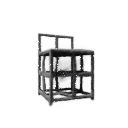}
        \end{subfigure}
        \begin{subfigure}{0.13\linewidth}
            \centering
            \includegraphics[width=\textwidth, trim=15pt 15pt 15pt 15pt, clip]{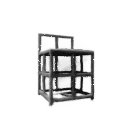}
        \end{subfigure}
        \begin{subfigure}{0.13\linewidth}
            \centering
            \includegraphics[width=\textwidth, trim=15pt 15pt 15pt 15pt, clip]{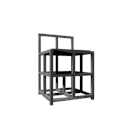}
        \end{subfigure}
    
        \vspace{-4pt}
    
        \begin{subfigure}{0.13\linewidth}
            \centering
            \includegraphics[width=\textwidth, trim=15pt 15pt 15pt 15pt, clip]{render_placeholder.png}
        \end{subfigure}
        \begin{subfigure}{0.13\linewidth}
            \centering
            \includegraphics[width=\textwidth, trim=15pt 15pt 15pt 15pt, clip]{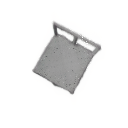}
        \end{subfigure}
        \begin{subfigure}{0.13\linewidth}
            \centering
            \includegraphics[width=\textwidth, trim=15pt 15pt 15pt 15pt, clip]{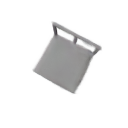}
        \end{subfigure}
        \begin{subfigure}{0.13\linewidth}
            \centering
            \includegraphics[width=\textwidth, trim=15pt 15pt 15pt 15pt, clip]{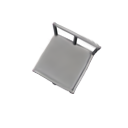}
        \end{subfigure}
        \begin{subfigure}{0.13\linewidth}
            \centering
            \includegraphics[width=\textwidth, trim=15pt 15pt 15pt 15pt, clip]{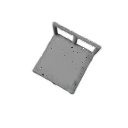}
        \end{subfigure}
        \begin{subfigure}{0.13\linewidth}
            \centering
            \includegraphics[width=\textwidth, trim=15pt 15pt 15pt 15pt, clip]{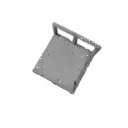}
        \end{subfigure}
        \begin{subfigure}{0.13\linewidth}
            \centering
            \includegraphics[width=\textwidth, trim=15pt 15pt 15pt 15pt, clip]{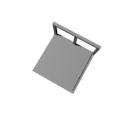}
        \end{subfigure}
    
        \vspace{-2pt}

        \begin{subfigure}{0.13\linewidth}
            \centering
            \includegraphics[width=\textwidth, trim=15pt 15pt 15pt 15pt, clip]{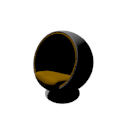}
        \end{subfigure}
        \begin{subfigure}{0.13\linewidth}
            \centering
            \includegraphics[width=\textwidth, trim=15pt 15pt 15pt 15pt, clip]{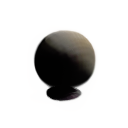}
        \end{subfigure}
        \begin{subfigure}{0.13\linewidth}
            \centering
            \includegraphics[width=\textwidth, trim=15pt 15pt 15pt 15pt, clip]{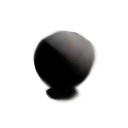}
        \end{subfigure}
        \begin{subfigure}{0.13\linewidth}
            \centering
            \includegraphics[width=\textwidth, trim=15pt 15pt 15pt 15pt, clip]{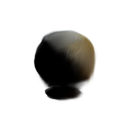}
        \end{subfigure}
        \begin{subfigure}{0.13\linewidth}
            \centering
            \includegraphics[width=\textwidth, trim=15pt 15pt 15pt 15pt, clip]{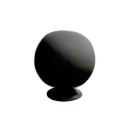}
        \end{subfigure}
        \begin{subfigure}{0.13\linewidth}
            \centering
            \includegraphics[width=\textwidth, trim=15pt 15pt 15pt 15pt, clip]{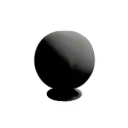}
        \end{subfigure}
        \begin{subfigure}{0.13\linewidth}
            \centering
            \includegraphics[width=\textwidth, trim=15pt 15pt 15pt 15pt, clip]{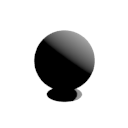}
        \end{subfigure}
    
        \vspace{-4pt}
    
        \begin{subfigure}{0.13\linewidth}
            \centering
            \includegraphics[width=\textwidth, trim=15pt 15pt 15pt 15pt, clip]{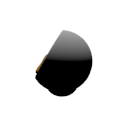}
        \end{subfigure}
        \begin{subfigure}{0.13\linewidth}
            \centering
            \includegraphics[width=\textwidth, trim=15pt 15pt 15pt 15pt, clip]{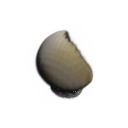}
        \end{subfigure}
        \begin{subfigure}{0.13\linewidth}
            \centering
            \includegraphics[width=\textwidth, trim=15pt 15pt 15pt 15pt, clip]{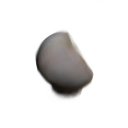}
        \end{subfigure}
        \begin{subfigure}{0.13\linewidth}
            \centering
            \includegraphics[width=\textwidth, trim=15pt 15pt 15pt 15pt, clip]{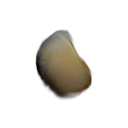}
        \end{subfigure}
        \begin{subfigure}{0.13\linewidth}
            \centering
            \includegraphics[width=\textwidth, trim=15pt 15pt 15pt 15pt, clip]{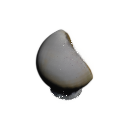}
        \end{subfigure}
        \begin{subfigure}{0.13\linewidth}
            \centering
            \includegraphics[width=\textwidth, trim=15pt 15pt 15pt 15pt, clip]{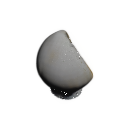}
        \end{subfigure}
        \begin{subfigure}{0.13\linewidth}
            \centering
            \includegraphics[width=\textwidth, trim=15pt 15pt 15pt 15pt, clip]{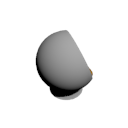}
        \end{subfigure}
    
        \vspace{-4pt}
    
        \begin{subfigure}{0.13\linewidth}
            \centering
            \includegraphics[width=\textwidth, trim=15pt 15pt 15pt 15pt, clip]{render_placeholder.png}
        \end{subfigure}
        \begin{subfigure}{0.13\linewidth}
            \centering
            \includegraphics[width=\textwidth, trim=15pt 15pt 15pt 15pt, clip]{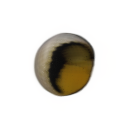}
        \end{subfigure}
        \begin{subfigure}{0.13\linewidth}
            \centering
            \includegraphics[width=\textwidth, trim=15pt 15pt 15pt 15pt, clip]{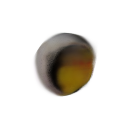}
        \end{subfigure}
        \begin{subfigure}{0.13\linewidth}
            \centering
            \includegraphics[width=\textwidth, trim=15pt 15pt 15pt 15pt, clip]{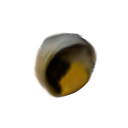}
        \end{subfigure}
        \begin{subfigure}{0.13\linewidth}
            \centering
            \includegraphics[width=\textwidth, trim=15pt 15pt 15pt 15pt, clip]{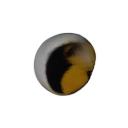}
        \end{subfigure}
        \begin{subfigure}{0.13\linewidth}
            \centering
            \includegraphics[width=\textwidth, trim=15pt 15pt 15pt 15pt, clip]{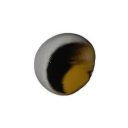}
        \end{subfigure}
        \begin{subfigure}{0.13\linewidth}
            \centering
            \includegraphics[width=\textwidth, trim=15pt 15pt 15pt 15pt, clip]{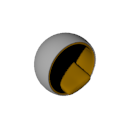}
        \end{subfigure}
    
        \vspace{-4pt}
    
        \begin{subfigure}{0.13\linewidth}
            \centering
            \includegraphics[width=\textwidth, trim=15pt 15pt 15pt 15pt, clip]{render_placeholder.png}
        \end{subfigure}
        \begin{subfigure}{0.13\linewidth}
            \centering
            \includegraphics[width=\textwidth, trim=15pt 15pt 15pt 15pt, clip]{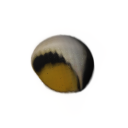}
        \end{subfigure}
        \begin{subfigure}{0.13\linewidth}
            \centering
            \includegraphics[width=\textwidth, trim=15pt 15pt 15pt 15pt, clip]{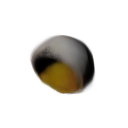}
        \end{subfigure}
        \begin{subfigure}{0.13\linewidth}
            \centering
            \includegraphics[width=\textwidth, trim=15pt 15pt 15pt 15pt, clip]{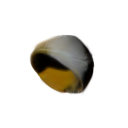}
        \end{subfigure}
        \begin{subfigure}{0.13\linewidth}
            \centering
            \includegraphics[width=\textwidth, trim=15pt 15pt 15pt 15pt, clip]{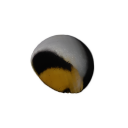}
        \end{subfigure}
        \begin{subfigure}{0.13\linewidth}
            \centering
            \includegraphics[width=\textwidth, trim=15pt 15pt 15pt 15pt, clip]{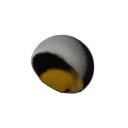}
        \end{subfigure}
        \begin{subfigure}{0.13\linewidth}
            \centering
            \includegraphics[width=\textwidth, trim=15pt 15pt 15pt 15pt, clip]{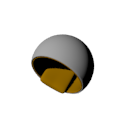}
        \end{subfigure}
    
        \caption{Additional Shapenet-SRN Chairs qualitative results on chairs in the testing split. Given two input views (one in the OpenLRM case), render novel views from around the object. Please zoom in to observe finer details.}
        \label{fig:supplement_chairs_results}
    \end{minipage}
    \end{figure*}

%% file: supplement_results.tex
\begin{figure*}[!t]
    \centering
    \begin{minipage}{0.75\linewidth}\centering
        \begin{subfigure}{0.13\linewidth}
            \centering
            Input View Pair
        \end{subfigure}
        \begin{subfigure}{0.13\linewidth}
            \centering
            OpenLRM
        \end{subfigure}
        \begin{subfigure}{0.13\linewidth}
            \centering
            pixelNeRF
        \end{subfigure}
        \begin{subfigure}{0.13\linewidth}
            \centering
            Splatter Image\\
            (1 view)
        \end{subfigure}
        \begin{subfigure}{0.13\linewidth}
            \centering
            Ours w/o bias
        \end{subfigure}
        \begin{subfigure}{0.13\linewidth}
            \centering
            Ours
        \end{subfigure}
        \begin{subfigure}{0.13\linewidth}
            \centering
            Ground Truth
        \end{subfigure}\vfill    

        \begin{subfigure}{0.13\linewidth}
            \centering
            \includegraphics[width=\textwidth, trim=15pt 15pt 15pt 15pt, clip]{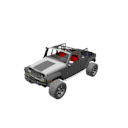}
        \end{subfigure}
        \begin{subfigure}{0.13\linewidth}
            \centering
            \includegraphics[width=\textwidth, trim=15pt 15pt 15pt 15pt, clip]{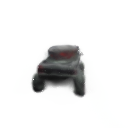}
        \end{subfigure}
        \begin{subfigure}{0.13\linewidth}
            \centering
            \includegraphics[width=\textwidth, trim=15pt 15pt 15pt 15pt, clip]{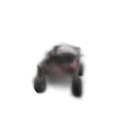}
        \end{subfigure}
        \begin{subfigure}{0.13\linewidth}
            \centering
            \includegraphics[width=\textwidth, trim=15pt 15pt 15pt 15pt, clip]{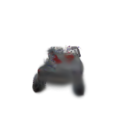}
        \end{subfigure}
        \begin{subfigure}{0.13\linewidth}
            \centering
            \includegraphics[width=\textwidth, trim=15pt 15pt 15pt 15pt, clip]{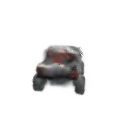}
        \end{subfigure}
        \begin{subfigure}{0.13\linewidth}
            \centering
            \includegraphics[width=\textwidth, trim=15pt 15pt 15pt 15pt, clip]{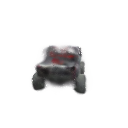}
        \end{subfigure}
        \begin{subfigure}{0.13\linewidth}
            \centering
            \includegraphics[width=\textwidth, trim=15pt 15pt 15pt 15pt, clip]{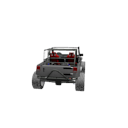}
        \end{subfigure}
    
        \vspace{-5pt}
    
        \begin{subfigure}{0.13\linewidth}
            \centering
            \includegraphics[width=\textwidth, trim=15pt 15pt 15pt 15pt, clip]{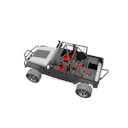}
        \end{subfigure}
        \begin{subfigure}{0.13\linewidth}
            \centering
            \includegraphics[width=\textwidth, trim=15pt 15pt 15pt 15pt, clip]{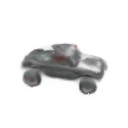}
        \end{subfigure}
        \begin{subfigure}{0.13\linewidth}
            \centering
            \includegraphics[width=\textwidth, trim=15pt 15pt 15pt 15pt, clip]{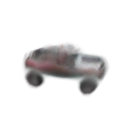}
        \end{subfigure}
        \begin{subfigure}{0.13\linewidth}
            \centering
            \includegraphics[width=\textwidth, trim=15pt 15pt 15pt 15pt, clip]{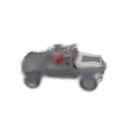}
        \end{subfigure}
        \begin{subfigure}{0.13\linewidth}
            \centering
            \includegraphics[width=\textwidth, trim=15pt 15pt 15pt 15pt, clip]{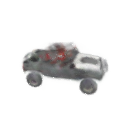}
        \end{subfigure}
        \begin{subfigure}{0.13\linewidth}
            \centering
            \includegraphics[width=\textwidth, trim=15pt 15pt 15pt 15pt, clip]{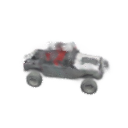}
        \end{subfigure}
        \begin{subfigure}{0.13\linewidth}
            \centering
            \includegraphics[width=\textwidth, trim=15pt 15pt 15pt 15pt, clip]{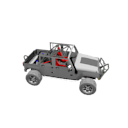}
        \end{subfigure}
    
        \vspace{-5pt}
    
        \begin{subfigure}{0.13\linewidth}
            \centering
            \includegraphics[width=\textwidth, trim=15pt 15pt 15pt 15pt, clip]{render_placeholder.png}
        \end{subfigure}
        \begin{subfigure}{0.13\linewidth}
            \centering
            \includegraphics[width=\textwidth, trim=15pt 15pt 15pt 15pt, clip]{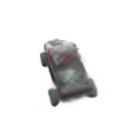}
        \end{subfigure}
        \begin{subfigure}{0.13\linewidth}
            \centering
            \includegraphics[width=\textwidth, trim=15pt 15pt 15pt 15pt, clip]{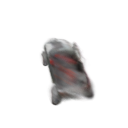}
        \end{subfigure}
        \begin{subfigure}{0.13\linewidth}
            \centering
            \includegraphics[width=\textwidth, trim=15pt 15pt 15pt 15pt, clip]{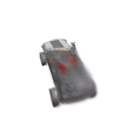}
        \end{subfigure}
        \begin{subfigure}{0.13\linewidth}
            \centering
            \includegraphics[width=\textwidth, trim=15pt 15pt 15pt 15pt, clip]{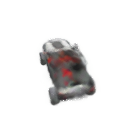}
        \end{subfigure}
        \begin{subfigure}{0.13\linewidth}
            \centering
            \includegraphics[width=\textwidth, trim=15pt 15pt 15pt 15pt, clip]{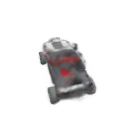}
        \end{subfigure}
        \begin{subfigure}{0.13\linewidth}
            \centering
            \includegraphics[width=\textwidth, trim=15pt 15pt 15pt 15pt, clip]{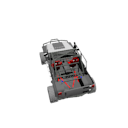}
        \end{subfigure}
    
        \vspace{-5pt}
    
        \begin{subfigure}{0.13\linewidth}
            \centering
            \includegraphics[width=\textwidth, trim=15pt 15pt 15pt 15pt, clip]{render_placeholder.png}
        \end{subfigure}
        \begin{subfigure}{0.13\linewidth}
            \centering
            \includegraphics[width=\textwidth, trim=15pt 15pt 15pt 15pt, clip]{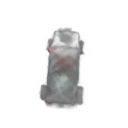}
        \end{subfigure}
        \begin{subfigure}{0.13\linewidth}
            \centering
            \includegraphics[width=\textwidth, trim=15pt 15pt 15pt 15pt, clip]{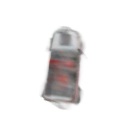}
        \end{subfigure}
        \begin{subfigure}{0.13\linewidth}
            \centering
            \includegraphics[width=\textwidth, trim=15pt 15pt 15pt 15pt, clip]{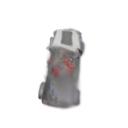}
        \end{subfigure}
        \begin{subfigure}{0.13\linewidth}
            \centering
            \includegraphics[width=\textwidth, trim=15pt 15pt 15pt 15pt, clip]{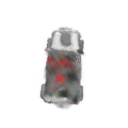}
        \end{subfigure}
        \begin{subfigure}{0.13\linewidth}
            \centering
            \includegraphics[width=\textwidth, trim=15pt 15pt 15pt 15pt, clip]{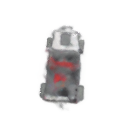}
        \end{subfigure}
        \begin{subfigure}{0.13\linewidth}
            \centering
            \includegraphics[width=\textwidth, trim=15pt 15pt 15pt 15pt, clip]{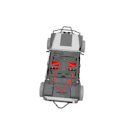}
        \end{subfigure}
    
        \vspace{-3pt}

        \begin{subfigure}{0.13\linewidth}
            \centering
            \includegraphics[width=\textwidth, trim=15pt 15pt 15pt 15pt, clip]{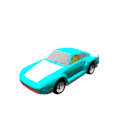}
        \end{subfigure}
        \begin{subfigure}{0.13\linewidth}
            \centering
            \includegraphics[width=\textwidth, trim=15pt 15pt 15pt 15pt, clip]{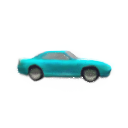}
        \end{subfigure}
        \begin{subfigure}{0.13\linewidth}
            \centering
            \includegraphics[width=\textwidth, trim=15pt 15pt 15pt 15pt, clip]{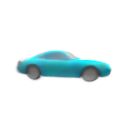}
        \end{subfigure}
        \begin{subfigure}{0.13\linewidth}
            \centering
            \includegraphics[width=\textwidth, trim=15pt 15pt 15pt 15pt, clip]{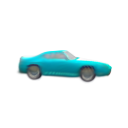}
        \end{subfigure}
        \begin{subfigure}{0.13\linewidth}
            \centering
            \includegraphics[width=\textwidth, trim=15pt 15pt 15pt 15pt, clip]{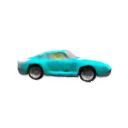}
        \end{subfigure}
        \begin{subfigure}{0.13\linewidth}
            \centering
            \includegraphics[width=\textwidth, trim=15pt 15pt 15pt 15pt, clip]{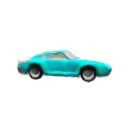}
        \end{subfigure}
        \begin{subfigure}{0.13\linewidth}
            \centering
            \includegraphics[width=\textwidth, trim=15pt 15pt 15pt 15pt, clip]{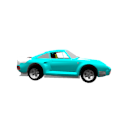}
        \end{subfigure}
    
        \vspace{-5pt}
    
        \begin{subfigure}{0.13\linewidth}
            \centering
            \includegraphics[width=\textwidth, trim=15pt 15pt 15pt 15pt, clip]{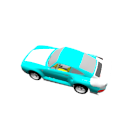}
        \end{subfigure}
        \begin{subfigure}{0.13\linewidth}
            \centering
            \includegraphics[width=\textwidth, trim=15pt 15pt 15pt 15pt, clip]{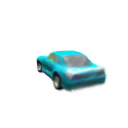}
        \end{subfigure}
        \begin{subfigure}{0.13\linewidth}
            \centering
            \includegraphics[width=\textwidth, trim=15pt 15pt 15pt 15pt, clip]{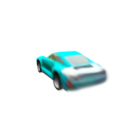}
        \end{subfigure}
        \begin{subfigure}{0.13\linewidth}
            \centering
            \includegraphics[width=\textwidth, trim=15pt 15pt 15pt 15pt, clip]{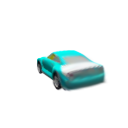}
        \end{subfigure}
        \begin{subfigure}{0.13\linewidth}
            \centering
            \includegraphics[width=\textwidth, trim=15pt 15pt 15pt 15pt, clip]{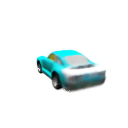}
        \end{subfigure}
        \begin{subfigure}{0.13\linewidth}
            \centering
            \includegraphics[width=\textwidth, trim=15pt 15pt 15pt 15pt, clip]{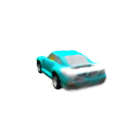}
        \end{subfigure}
        \begin{subfigure}{0.13\linewidth}
            \centering
            \includegraphics[width=\textwidth, trim=15pt 15pt 15pt 15pt, clip]{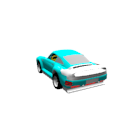}
        \end{subfigure}
    
        \vspace{-5pt}
    
        \begin{subfigure}{0.13\linewidth}
            \centering
            \includegraphics[width=\textwidth, trim=15pt 15pt 15pt 15pt, clip]{render_placeholder.png}
        \end{subfigure}
        \begin{subfigure}{0.13\linewidth}
            \centering
            \includegraphics[width=\textwidth, trim=15pt 15pt 15pt 15pt, clip]{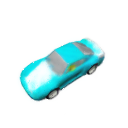}
        \end{subfigure}
        \begin{subfigure}{0.13\linewidth}
            \centering
            \includegraphics[width=\textwidth, trim=15pt 15pt 15pt 15pt, clip]{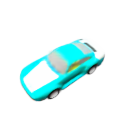}
        \end{subfigure}
        \begin{subfigure}{0.13\linewidth}
            \centering
            \includegraphics[width=\textwidth, trim=15pt 15pt 15pt 15pt, clip]{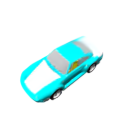}
        \end{subfigure}
        \begin{subfigure}{0.13\linewidth}
            \centering
            \includegraphics[width=\textwidth, trim=15pt 15pt 15pt 15pt, clip]{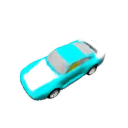}
        \end{subfigure}
        \begin{subfigure}{0.13\linewidth}
            \centering
            \includegraphics[width=\textwidth, trim=15pt 15pt 15pt 15pt, clip]{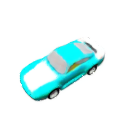}
        \end{subfigure}
        \begin{subfigure}{0.13\linewidth}
            \centering
            \includegraphics[width=\textwidth, trim=15pt 15pt 15pt 15pt, clip]{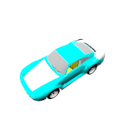}
        \end{subfigure}
    
        \vspace{-5pt}
    
        \begin{subfigure}{0.13\linewidth}
            \centering
            \includegraphics[width=\textwidth, trim=15pt 15pt 15pt 15pt, clip]{render_placeholder.png}
        \end{subfigure}
        \begin{subfigure}{0.13\linewidth}
            \centering
            \includegraphics[width=\textwidth, trim=15pt 15pt 15pt 15pt, clip]{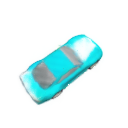}
        \end{subfigure}
        \begin{subfigure}{0.13\linewidth}
            \centering
            \includegraphics[width=\textwidth, trim=15pt 15pt 15pt 15pt, clip]{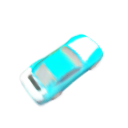}
        \end{subfigure}
        \begin{subfigure}{0.13\linewidth}
            \centering
            \includegraphics[width=\textwidth, trim=15pt 15pt 15pt 15pt, clip]{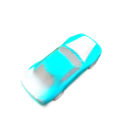}
        \end{subfigure}
        \begin{subfigure}{0.13\linewidth}
            \centering
            \includegraphics[width=\textwidth, trim=15pt 15pt 15pt 15pt, clip]{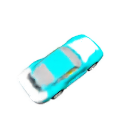}
        \end{subfigure}
        \begin{subfigure}{0.13\linewidth}
            \centering
            \includegraphics[width=\textwidth, trim=15pt 15pt 15pt 15pt, clip]{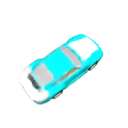}
        \end{subfigure}
        \begin{subfigure}{0.13\linewidth}
            \centering
            \includegraphics[width=\textwidth, trim=15pt 15pt 15pt 15pt, clip]{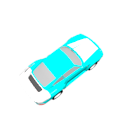}
        \end{subfigure}
    
        \vspace{-3pt}

        \begin{subfigure}{0.13\linewidth}
            \centering
            \includegraphics[width=\textwidth, trim=15pt 15pt 15pt 15pt, clip]{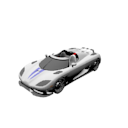}
        \end{subfigure}
        \begin{subfigure}{0.13\linewidth}
            \centering
            \includegraphics[width=\textwidth, trim=15pt 15pt 15pt 15pt, clip]{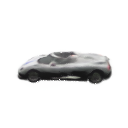}
        \end{subfigure}
        \begin{subfigure}{0.13\linewidth}
            \centering
            \includegraphics[width=\textwidth, trim=15pt 15pt 15pt 15pt, clip]{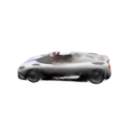}
        \end{subfigure}
        \begin{subfigure}{0.13\linewidth}
            \centering
            \includegraphics[width=\textwidth, trim=15pt 15pt 15pt 15pt, clip]{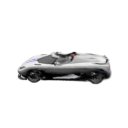}
        \end{subfigure}
        \begin{subfigure}{0.13\linewidth}
            \centering
            \includegraphics[width=\textwidth, trim=15pt 15pt 15pt 15pt, clip]{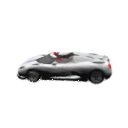}
        \end{subfigure}
        \begin{subfigure}{0.13\linewidth}
            \centering
            \includegraphics[width=\textwidth, trim=15pt 15pt 15pt 15pt, clip]{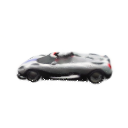}
        \end{subfigure}
        \begin{subfigure}{0.13\linewidth}
            \centering
            \includegraphics[width=\textwidth, trim=15pt 15pt 15pt 15pt, clip]{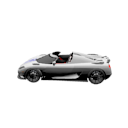}
        \end{subfigure}
    
        \vspace{-5pt}
    
        \begin{subfigure}{0.13\linewidth}
            \centering
            \includegraphics[width=\textwidth, trim=15pt 15pt 15pt 15pt, clip]{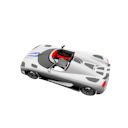}
        \end{subfigure}
        \begin{subfigure}{0.13\linewidth}
            \centering
            \includegraphics[width=\textwidth, trim=15pt 15pt 15pt 15pt, clip]{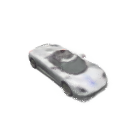}
        \end{subfigure}
        \begin{subfigure}{0.13\linewidth}
            \centering
            \includegraphics[width=\textwidth, trim=15pt 15pt 15pt 15pt, clip]{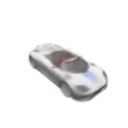}
        \end{subfigure}
        \begin{subfigure}{0.13\linewidth}
            \centering
            \includegraphics[width=\textwidth, trim=15pt 15pt 15pt 15pt, clip]{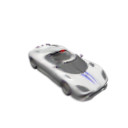}
        \end{subfigure}
        \begin{subfigure}{0.13\linewidth}
            \centering
            \includegraphics[width=\textwidth, trim=15pt 15pt 15pt 15pt, clip]{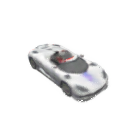}
        \end{subfigure}
        \begin{subfigure}{0.13\linewidth}
            \centering
            \includegraphics[width=\textwidth, trim=15pt 15pt 15pt 15pt, clip]{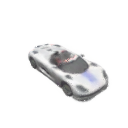}
        \end{subfigure}
        \begin{subfigure}{0.13\linewidth}
            \centering
            \includegraphics[width=\textwidth, trim=15pt 15pt 15pt 15pt, clip]{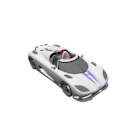}
        \end{subfigure}
    
        \vspace{-5pt}
    
        \begin{subfigure}{0.13\linewidth}
            \centering
            \includegraphics[width=\textwidth, trim=15pt 15pt 15pt 15pt, clip]{render_placeholder.png}
        \end{subfigure}
        \begin{subfigure}{0.13\linewidth}
            \centering
            \includegraphics[width=\textwidth, trim=15pt 15pt 15pt 15pt, clip]{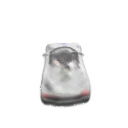}
        \end{subfigure}
        \begin{subfigure}{0.13\linewidth}
            \centering
            \includegraphics[width=\textwidth, trim=15pt 15pt 15pt 15pt, clip]{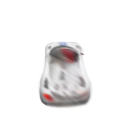}
        \end{subfigure}
        \begin{subfigure}{0.13\linewidth}
            \centering
            \includegraphics[width=\textwidth, trim=15pt 15pt 15pt 15pt, clip]{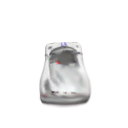}
        \end{subfigure}
        \begin{subfigure}{0.13\linewidth}
            \centering
            \includegraphics[width=\textwidth, trim=15pt 15pt 15pt 15pt, clip]{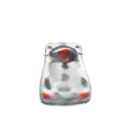}
        \end{subfigure}
        \begin{subfigure}{0.13\linewidth}
            \centering
            \includegraphics[width=\textwidth, trim=15pt 15pt 15pt 15pt, clip]{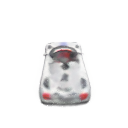}
        \end{subfigure}
        \begin{subfigure}{0.13\linewidth}
            \centering
            \includegraphics[width=\textwidth, trim=15pt 15pt 15pt 15pt, clip]{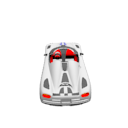}
        \end{subfigure}
    
        \begin{subfigure}{0.13\linewidth}
            \centering
            \includegraphics[width=\textwidth, trim=15pt 15pt 15pt 15pt, clip]{render_placeholder.png}
        \end{subfigure}
        \begin{subfigure}{0.13\linewidth}
            \centering
            \includegraphics[width=\textwidth, trim=15pt 15pt 15pt 15pt, clip]{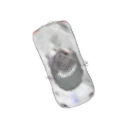}
        \end{subfigure}
        \begin{subfigure}{0.13\linewidth}
            \centering
            \includegraphics[width=\textwidth, trim=15pt 15pt 15pt 15pt, clip]{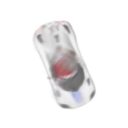}
        \end{subfigure}
        \begin{subfigure}{0.13\linewidth}
            \centering
            \includegraphics[width=\textwidth, trim=15pt 15pt 15pt 15pt, clip]{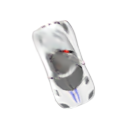}
        \end{subfigure}
        \begin{subfigure}{0.13\linewidth}
            \centering
            \includegraphics[width=\textwidth, trim=15pt 15pt 15pt 15pt, clip]{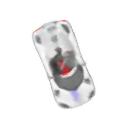}
        \end{subfigure}
        \begin{subfigure}{0.13\linewidth}
            \centering
            \includegraphics[width=\textwidth, trim=15pt 15pt 15pt 15pt, clip]{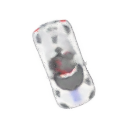}
        \end{subfigure}
        \begin{subfigure}{0.13\linewidth}
            \centering
            \includegraphics[width=\textwidth, trim=15pt 15pt 15pt 15pt, clip]{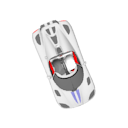}
        \end{subfigure}
        \caption{Additional Shapenet-SRN Cars qualitative results on cars in the testing split. Given two input views (one in the OpenLRM and Splatter Image case), render novel views from around the object. Please zoom in to observe finer details.}
        \label{fig:supplement_results}
    \end{minipage}
    \end{figure*}